\documentclass[preprint,12pt]{elsarticle}

\usepackage{graphicx}
\usepackage{bm}
\usepackage{amsmath}
\usepackage{setspace}
\usepackage{hyperref}
\hypersetup{
    colorlinks=true,       % false: boxed links; true: colored links
}
\usepackage{amssymb}
\usepackage{lineno}

\journal{Journal Name}

\begin{document}

\begin{frontmatter}


\title{Numerical Gaussian Processes\\
for Time-dependent and Non-linear\\
Partial Differential Equations}



\author{Maziar Raissi$^{1}$, Paris Perdikaris$^{2}$, and George Em Karniadakis$^{1}$}

\address{$^{1}$Division of Applied Mathematics, Brown University,\\ Providence, RI, 02912, USA\\
$^{2}$Department of Mechanical Engineering,\\ Massachusetts Institute of Technology,\\ Cambridge, MA, 02139, USA}

% Points to make:
% -  The differential operator and the choice  of time-stepping scheme define a structured GP prior.
% - Time-stepping and uncertainty propagation in  PDEs (contrast with existing reverse-engineered methods for simple ODE systems)
% - this solves a formidable infinite dimensional UQ problem
% - General: any equation, any time-stepping scheme: multi-step, multi-stage (explicit and implicit)
% - Streaming forcing data, a-posteriori error estimates, etc.

\begin{abstract}
We introduce the concept of \emph{numerical Gaussian processes}, which we define as Gaussian processes with covariance functions resulting from temporal discretization of time-dependent partial differential equations. Numerical Gaussian processes, by construction, are designed to deal with cases where: (1) all we observe are noisy data on \emph{black-box} initial conditions, and (2) we are interested in \emph{quantifying the uncertainty} associated with such noisy data in our solutions to time-dependent partial differential equations. Our method circumvents the need for spatial discretization of the differential operators by proper placement of Gaussian process priors. This is an attempt to construct structured and data-efficient learning machines, which are explicitly informed by the underlying physics that possibly generated the observed data.  The effectiveness of the proposed approach is demonstrated through several benchmark problems involving linear and nonlinear time-dependent operators. In all examples, we are able to recover accurate approximations of the latent solutions, and consistently propagate uncertainty, even in cases involving very long time integration. 
% {\color{red}{We demonstrate the method by considering ... examples ... say something about accuracy, UQ, scalability to big domains, long-term integration.}}
\end{abstract}


\end{frontmatter}

% \linenumbers

\section{Introduction} \label{sec:methods}

Data-driven methods are taking center stage across many disciplines of science, and machine learning techniques have achieved groundbreaking results across a diverse spectrum of pattern recognition tasks \cite{krizhevsky2012imagenet, hochreiter1997long,ghahramani2015probabilistic,lecun2015deep,jordan2015machine}. Despite their disruptive implications, many of these methods are blind to any underlying laws of physics that may have shaped the distribution of the observed data. A natural question would then be how one can construct efficient learning machines that explicitly leverage such structured prior information? To answer this question we have to turn our attention to the immense collective knowledge originating from centuries of research in applied mathematics and mathematical physics. Modeling the physical world through the lens of mathematics typically translates into deriving conservation laws from first principles, which often take the form of systems of partial differential equations. In many practical settings, the solution of such systems is only accessible by means of numerical algorithms that provide sensible approximations to given quantities of interest. In this work, we aim to capitalize on the long-standing developments of classical methods in numerical analysis and revisit partial differential equations from a {\em statistical inference} viewpoint. The merits of this approach are twofold. First, it enables the construction of data-efficient learning machines that can encode physical conservation laws as structured prior information. Second, it allows the design of novel numerical algorithms that can seamlessly blend equations and noisy data, infer latent quantities of interest (e.g., the solution to a partial differential equation), and naturally quantify uncertainty in computations. This approach is aligned in spirit with the emerging field of probabilistic numerics \cite{probnum}, which roots all the way back to Poincar\'{e}'s courses on probability theory \cite{poincare1896}, and has been recently revived by the pioneering works of \cite{diaconis1988bayesian,o1992some,HenOsbGirRSPA2015,conrad2016statistical}.

To illustrate the key ingredients of this study, let us start by considering linear\footnote{Non-linear equations have to be studied on a case by case basis (see e.g., section \ref{sec:Burgers}).} partial differential equations of the form
\begin{eqnarray}\label{eq:Linear}
&&u_t = \mathcal{L}_x u,\ x \in \Omega, \ t\in[0,T],
\end{eqnarray}
where $\mathcal{L}_x$ is a linear operator and $u(t,x)$ denotes the latent solution. As an example, the one dimensional heat equation corresponds to the case where $\mathcal{L}_x = \frac{\partial^2}{\partial x^2}$. Moreover, $\Omega$ is a subset of $\mathbb{R}^D$. All we observe are noisy data $\{\bm{x}^0,\bm{u}^0\}$ on the \emph{black-box} initial function $u(0,x)$ as well as some information on the domain boundary $\partial \Omega$ to be specified later. Our goal is to predict the latent solution $u(t,x)$ at $t>0$, and propagate the uncertainty due to noise in the initial data. For starters, let us try to convey the main ideas of this work using the Euler time stepping scheme 
\begin{equation}\label{eq:ForwardEuler}
u^{n} = u^{n-1} + \Delta t\ \mathcal{L}_x u^{n-1}.
\end{equation}
Here, $u^n(x) = u(t^n,x)$. Building upon Raissi et al. \cite{Raissi2017736, raissi2017machine}, we place a Gaussian process \cite{Rasmussen06gaussianprocesses} prior on $u^{n-1}$, i.e.,
\begin{equation}\label{eq:GP_prior_for_Euler}
u^{n-1}(x) \sim \mathcal{GP}(0, k^{n-1,n-1}_{u,u}(x,x',\theta)).
\end{equation}
Here, $\theta$ denotes the hyper-parameters of the covariance function $k_{u,u}^{n-1,n-1}$. Gaussian process regression (see \cite{Rasmussen06gaussianprocesses, murphy2012machine}) is a non-parametric Bayesian machine learning technique that provides a flexible prior distribution over functions, enjoys analytical tractability, and has a fully probabilistic work-flow that returns robust posterior variance estimates, which quantify uncertainty in a natural way. Moreover, Gaussian processes are among a class of methods known as kernel machines (see \cite{vapnik2013nature, scholkopf2002learning, tipping2001sparse}) and are analogous to regularization approaches (see \cite{tikhonov1963solution, Tikhonov/Arsenin/77, poggio1990networks}). They can also be viewed as a prior on one-layer feed-forward Bayesian neural networks with an infinite number of hidden units \cite{neal2012bayesian}. The Gaussian process prior assumption (\ref{eq:GP_prior_for_Euler}) along with the Euler scheme (\ref{eq:ForwardEuler}) will allow us to capture the entire structure of the differential operator $\mathcal{L}_x$ as well as the Euler time-stepping rule in the resulting multi-output Gaussian process
\begin{equation}\label{eq:ForwardEulerGP}
\left[\begin{array}{c}
u^{n} \\ 
u^{n-1}
\end{array} \right] \sim \mathcal{GP}\left(0, \left[\begin{array}{cc}
k^{n,n}_{u,u} & k^{n,n-1}_{u,u}\\ 
 & k^{n-1,n-1}_{u,u}
\end{array} \right]\right).
\end{equation}
The specific forms of the kernels $k^{n,n}_{u,u}$ and $k^{n,n-1}_{u,u}$ are direct functions of the Euler scheme (\ref{eq:ForwardEuler}) as well as the prior assumption (\ref{eq:GP_prior_for_Euler}), and will be discussed in more detail later. The multi-output process (\ref{eq:ForwardEulerGP}) is an example of a \emph{numerical Gaussian process}, because the covariance functions $k^{n,n}_{u,u}$ and $k^{n,n-1}_{u,u}$ result from a numerical scheme, in this case, the Euler method. Essentially, this introduces a structured prior that explicitly encodes the physical law modeled by the partial differential equation (\ref{eq:Linear}). In the following, we will generalize the framework outlined above to  arbitrary \emph{linear multi-step methods}, originally proposed by Bashforth and Adams \cite{bashforth1883attempt}, as well as  \emph{Runge-Kutta methods}, generally attributed to Runge \cite{runge1895numerische}. The biggest challenge here is the proper placement of the Gaussian process prior (see e.g., equation (\ref{eq:GP_prior_for_Euler})) in order to avoid inversion of differential operators and to bypass the classical need for spatial discretization of such operators. For instance, in the above example (see equations (\ref{eq:ForwardEuler}) and (\ref{eq:GP_prior_for_Euler})), it would have been an inappropriate choice to start by placing a Gaussian process prior on $u^{n}$, rather than on $u^{n-1}$, as obtaining the \emph{numerical Gaussian process} (\ref{eq:ForwardEulerGP}) would then involve inverting operators of the form $I + \Delta t \mathcal{L}_x$ corresponding to the Euler method. Moreover, propagating the uncertainty associated with the noisy initial observations $\{\bm{x}^0, \bm{u}^0\}$ through time is another major challenge addressed in the following.

\section{Linear Multi-step Methods}\label{sec:LinearMultistepMethods}
Let us start with the most general form of the linear multi-step methods \cite{butcher2016numerical} applied to equation (\ref{eq:Linear}); i.e.,
\begin{equation}\label{eq:LinearMultistepMethods}
u^{n} = \sum_{i=1}^m \alpha_i u^{n-i} + \Delta t \sum_{i=0}^m \beta_i \mathcal{L}_x u^{n-i}.
\end{equation}
Different choices for the parameters $\alpha_i$ and $\beta_i$ result in specific schemes. For instance, in table \ref{table:ThreeExamples}, we present some specific members of the family of linear multi-step methods (\ref{eq:LinearMultistepMethods}).
\begin{table}
\caption{Some specific members of the family of linear multi-step methods (\ref{eq:LinearMultistepMethods}).}\label{table:ThreeExamples}
\onehalfspacing
\begin{center}
\begin{tabular}{|c|c|}
\hline
 Forward Euler & $u^{n} = u^{n-1} + \Delta t \mathcal{L}_x u^{n-1}$\\
 \hline
 \hline
 Backward Euler & $u^{n} = u^{n-1} + \Delta t \mathcal{L}_x u^{n}$\\ 
 \hline
 \hline
 Trapezoidal Rule & $u^{n} = u^{n-1} + \frac12 \Delta t \mathcal{L}_x u^{n-1} + \frac12 \Delta t \mathcal{L}_x u^{n}$\\  
 \hline
\end{tabular}
\end{center}
\end{table}
We encourage the reader to keep these special cases in mind while reading the rest of this section.  Linear multi-step methods (\ref{eq:LinearMultistepMethods}) can be equivalently written as
\begin{equation}\label{eq:LinearMultistepMethods_operator_form}
\mathcal{P}_x u^{n} = \sum_{i=1}^m \mathcal{Q}^i_x u^{n-i},
\end{equation}
where $\mathcal{P}_x u := u - \Delta t \beta_0 \mathcal{L}_x u$ and $\mathcal{Q}^i_x u := \alpha_i u + \Delta t \beta_i \mathcal{L}_x u$. Some special cases of equation (\ref{eq:LinearMultistepMethods_operator_form}) are given in table \ref{table:ThreeExamples_operator_form}.
\begin{table}
\caption{Some special cases of equation (\ref{eq:LinearMultistepMethods_operator_form}).}\label{table:ThreeExamples_operator_form}
\onehalfspacing
\begin{center}
\begin{tabular}{| c|c| }
\hline
 Forward Euler & $u^{n} =  \mathcal{Q}_x u^{n-1}$  \\
 & $\mathcal{Q}_x u^{n-1} = u^{n-1}+ \Delta t \mathcal{L}_x u^{n-1}$ \\
 \hline
 \hline
 Backward Euler & $\mathcal{P}_x u^{n} = u^{n-1}$  \\ 
& $\mathcal{P}_x u^{n} = u^{n} - \Delta t \mathcal{L}_x u^{n}$\\ 
 \hline
 \hline
 Trapezoidal Rule & $\mathcal{P}_x u^{n} = \mathcal{Q}_x u^{n-1}$ \\
 & $\mathcal{P}_x u^{n} = u^{n} - \frac12 \Delta t \mathcal{L}_x u^{n}$\\
 & $\mathcal{Q}_x u^{n-1} = u^{n-1} + \frac12 \Delta t \mathcal{L}_x u^{n-1}$ \\
 \hline
\end{tabular}
\end{center}
\end{table}
For every $j=0,1,\ldots,m$ and some $\tau \in [0,1]$ which depends on the specific choices for the values of the parameters $\alpha_i$ and $\beta_i$, we define $u^{n-j+\tau}$ to be given by
\begin{eqnarray}\label{eq:LinearMultistepMethods_half_way}
\mathcal{P}_x u^{n-j+1} =: u^{n-j+\tau} := \sum_{i=1}^m \mathcal{Q}^i_x u^{n-i-j+1}.
\end{eqnarray}
Definition (\ref{eq:LinearMultistepMethods_half_way}) takes the specific forms given in table \ref{table:ThreeExamples_half_way} for some example schemes.
\begin{table}
\caption{Some special cases of equation (\ref{eq:LinearMultistepMethods_half_way}).}\label{table:ThreeExamples_half_way}
\onehalfspacing
\begin{center}
\begin{tabular}{| c|c| c|}
\hline
 Forward Euler & $u^{n} =  \mathcal{Q}_x u^{n-1}$  \\
 & $\tau = 0$ \\
 \hline
 \hline
 Backward Euler & $\mathcal{P}_x u^{n} = u^{n-1}$  \\ 
& $\tau = 1$\\ 
 \hline
 \hline
 Trapezoidal Rule & $\mathcal{P}_x u^{n} = u^{n-1/2} = \mathcal{Q}_x u^{n-1}$ \\
 & $\tau = 1/2$ \\
 \hline
\end{tabular}
\end{center}
\end{table}
Shifting every term involved in the above definition (\ref{eq:LinearMultistepMethods_half_way}) by $-\tau$ yields
\begin{eqnarray}\label{eq:LinearMultistepMethods_shifted}
\mathcal{P}_x u^{n-j+1-\tau} = u^{n-j} = \sum_{i=1}^m \mathcal{Q}^i_x u^{n-i-j+1-\tau}.
\end{eqnarray}
To give an example, for the trapezoidal rule we obtain $\mathcal{P}_x u^{n+1/2} = u^{n} = \mathcal{Q}_x u^{n-1/2}$ and $\mathcal{P}_x u^{n-1/2} = u^{n-1} = \mathcal{Q}_x u^{n-3/2}$. Therefore, as a direct consequence of equation (\ref{eq:LinearMultistepMethods_shifted}) we have
\begin{eqnarray}
u^{n} &=& \sum_{i=1}^m \mathcal{Q}^i_x u^{n-i+1-\tau},\ \ \ \text{when}\ \ \  j = 0,\\
u^{n-j} &=& \mathcal{P}_x u^{n-j+1-\tau},\ \ \ \text{when}\ \ \ j=1,\ldots,m.\nonumber
\end{eqnarray}
This, in the special case of the trapezoidal rule, translates to $u^n = \mathcal{Q}_x u^{n-1/2}$ and $u^{n-1} = \mathcal{P}_x u^{n-1/2}$. It is worth noting that by assuming $u^{n-1/2}(x) \sim \mathcal{GP}(0,k(x,x';\theta))$, we can capture the entire structure of the trapezoidal rule in the resulting joint distribution of $u^n$ and $u^{n-1}$. This proper placement of the Gaussian process prior is key to the proposed methodology as it allows us to avoid any spatial discretization of differential operators since no inversion of such operators is necessary. We will capitalize on this idea in the following.

\subsection{Prior}
Assuming that
\begin{equation}\label{LinearMultistepMethods_prior_assumption}
u^{n-j+1-\tau}(x) \sim \mathcal{GP}(0,k^{j,j}(x,x';\theta_j)),\ \ \ j=1,\ldots,m,
\end{equation}
are $m$ independent processes, we obtain the following \emph{numerical Gaussian process}
\[
\left[\begin{array}{c}
u^{n} \\
\vdots \\
u^{n-m}
\end{array} \right] \sim \mathcal{GP}\left(0,\left[\begin{array}{ccc}
k^{n,n}_{u,u} &  \cdots & k^{n,n-m}_{u,u}\\ 
 & \ddots & \vdots \\
 &  & k^{n-m,n-m}_{u,u}
\end{array} \right]\right),
\]
where
\begin{equation}\label{eq:LinearMultistep_Kernels}
\begin{array}{ll}
k^{n,n}_{u,u} = \sum_{i=1}^m \mathcal{Q}^i_x\mathcal{Q}^i_{x'} k^{i,i}, & k^{n,n-j}_{u,u} = \mathcal{Q}^j_x \mathcal{P}_{x'} k^{j,j},\\
k^{n-i,n-j}_{u,u} = 0, \ \ \ i \neq j, & k^{n-j,n-j}_{u,u} = \mathcal{P}_x\mathcal{P}_{x'} k^{j,j}, \ \ \ j=1,\ldots,m.
\end{array}
\end{equation}
It is worth noting that the entire structure of linear multi-step methods (\ref{eq:LinearMultistepMethods}) is captured by the kernels given in equations (\ref{eq:LinearMultistep_Kernels}). Note that although we start from an independence   assumption in equation (\ref{LinearMultistepMethods_prior_assumption}), the resulting \emph{numerical Gaussian process} exhibits a fully correlated structure as illustrated in equations (\ref{eq:LinearMultistep_Kernels}). Moreover, the information on the boundary $\partial \Omega$ of the domain $\Omega$ can often be summarized by noisy observations $\{\bm{x}^n_b,\bm{u}^n_b\}$ of a linear transformation $\mathcal{B}_x$ of $u^n$; i.e., noisy data on
\[
u^n_b := \mathcal{B}_x u^n.
\]
Using this, we obtain the following covariance functions involving the boundary
\begin{eqnarray*}
k^{n,n}_{b,u} = \mathcal{B}_{x}k^{n,n}_{u,u}, & k^{n,n}_{b,b} = \mathcal{B}_{x}\mathcal{B}_{x'}k^{n,n}_{u,u}, & k^{n,n-j}_{b,u} = \mathcal{B}_{x}k^{n,n-j}_{u,u},\ \ \ j=1,\ldots,m.
\end{eqnarray*}
The numerical examples accompanying this manuscript are designed to showcase different special treatments of boundary conditions, including  Dirichlet, Neumann, mixed, and periodic boundary conditions.

\subsection{Work flow and computational cost}\label{sec:Workflow}
The proposed work flow is summarized below:
\begin{enumerate}
\item Starting from the initial data $\{\bm{x}^{0},\bm{u}^{0}\}$ and the boundary data $\{\bm{x}^{1}_b, \bm{u}^{1}_b\}$, we train the kernel hyper-parameters as outlined in section \ref{sec:Traning}. This step carries the main computational burden as it scales cubically with the total number of training points since it involves  Cholesky factorization of full symmetric positive-definite covariance matrices \cite{Rasmussen06gaussianprocesses}.
\item Having identified the optimal set of kernel hyper-parameters, we utilize the conditional posterior distribution to predict the solution at the next time-step and generate the  \emph{artificial data} $\{\bm{x}^{1},\bm{u}^{1}\}$. Note that $\bm{x}^{1}$ is randomly sampled in the spatial domain according to a uniform distribution, and  $\bm{u}^{1}$ is a normally distributed random vector, as outlined in section \ref{sec:Prediction}.
\item Given the \emph{artificial data} $\{\bm{x}^{1},\bm{u}^{1}\}$ and boundary data $\{\bm{x}^{2}_b, \bm{u}^{2}_b\}$ we proceed with training the kernel hyper-parameters for the second time-step\footnote{To be precise, we are using the mean of the random vector $\bm{u}^{1}$ for training purposes.} (see section \ref{sec:Traning}).
\item Having identified the optimal set of kernel hyper-parameters, we utilize the conditional posterior distribution to predict the solution at the next time-step and generate the  \emph{artificial data} $\{\bm{x}^{2},\bm{u}^{2}\}$, where $\bm{x}^{2}$ is randomly sampled in the spatial domain according to a uniform distribution. However, since  $\bm{u}^{1}$ is a random vector, we have to marginalize it out in order to obtain consistent uncertainty estimates for $\bm{u}^{2}$. This procedure is outlined in section \ref{sec:Uncertainty}.
\item Steps 3 and 4 are repeated until the final integration time is reached.
\end{enumerate}

In summary, the proposed methodology boils down to a sequence of Gaussian process regressions at every time-step. To accelerate training, one can use the optimal set of hyper-parameters from the previous time-step as an initial guess for the current one.

\subsection{Training}\label{sec:Traning}
In the following, for notational convenience and without loss of generality\footnote{The reader should be able to figure out the details without much difficulty while generalizing to cases with $m>1$. Moreover, for the examples accompanying this manuscript, more details are also provided in the appendix.}, we will operate under the assumption that $m=1$ (see equation (\ref{eq:LinearMultistepMethods})). The hyper-parameters $\theta_i,\ i=1,\ldots,m$, can be trained by employing the Negative Log Marginal Likelihood resulting from
\begin{equation}\label{eq:LinearMultistepMethods_NLML}
\left[\begin{array}{c}
\bm{u}^{n}_b \\ 
\bm{u}^{n-1}
\end{array} \right] \sim \mathcal{N}\left(0,\bm{K}\right),
\end{equation}
where $\{\bm{x}^{n}_b, \bm{u}^{n}_b\}$ are the (noisy) data on the boundary, $\{\bm{x}^{n-1}, \bm{u}^{n-1}\}$  are \emph{artificially generated data} to be explained later (see equation (\ref{eq:LinearMultistepMethods_artificial_data})), and
\[
\bm{K} := \left[\begin{array}{cc}
k^{n,n}_{b,b}(\bm{x}_b^{n},\bm{x}_b^{n}) + \sigma_{n}^2 I & k^{n,n-1}_{b,u}(\bm{x}_b^{n},\bm{x}^{n-1})\\ 
 & k^{n-1,n-1}_{u,u}(\bm{x}^{n-1},\bm{x}^{n-1}) + \sigma^2_{n-1}I
\end{array} \right].
\]
It is worth mentioning that the marginal likelihood provides a natural regularization mechanism that balances the trade-off between data fit and model complexity. This effect is known as Occam's razor \cite{rasmussen2001occam} after William of Occam 1285--1349 who encouraged simplicity in explanations by the principle: ``plurality should not be assumed without necessity". 

\subsection{Posterior}\label{sec:Prediction}
In order to predict $u^{n}(x^{n}_*)$ at a new test point $x^{n}_*$, we use the following conditional distribution
\begin{eqnarray*}
u^{n}(x^{n}_*)\ | \left[\begin{array}{c}
\bm{u}^{n}_b \\ 
\bm{u}^{n-1}
\end{array} \right] \sim \mathcal{N}\left(
\bm{q}^T
 \bm{K}^{-1}\left[\begin{array}{c}
\bm{u}^{n}_b \\ 
\bm{u}^{n-1}
\end{array} \right], k^{n,n}_{u,u}(x^{n}_*,x^{n}_*)  -  \bm{q}^T\bm{K}^{-1}\bm{q}\right),
\end{eqnarray*}
where
\[
\bm{q}^T := \left[\begin{array}{cc}
k^{n,n}_{u,b}(x_*^{n},\bm{x}^{n}_b)  & k^{n,n-1}_{u,u}(x_*^{n},\bm{x}^{n-1})
\end{array} \right].
\]
\subsection{Propagating Uncertainty}\label{sec:Uncertainty}
However, to properly propagate the uncertainty associated with the initial data through time, one should not stop here. Since $\{\bm{x}^{n-1}, \bm{u}^{n-1}\}$ are \emph{artificially generated data} (see equation (\ref{eq:LinearMultistepMethods_artificial_data})) we have to marginalize them out by employing
\[
\bm{u}^{n-1} \sim \mathcal{N}\left(
\bm{\mu}^{n-1},\bm{\Sigma}^{n-1,n-1}\right),
\]
to obtain
\begin{eqnarray}\label{LinearMultistepMethods_posterior}
&&
u^{n}(x^{n}_*)\ |\ \bm{u}^{n}_b
\sim \mathcal{N}\left(\mu^{n}(x_*^{n}),  \Sigma^{n,n}(x_*^{n},x_*^{n})\right),
\end{eqnarray}
where
\[
\mu^{n}(x_*^{n}) = \bm{q}^T \bm{K}^{-1}\left[\begin{array}{c}
\bm{u}^{n}_b \\ 
\bm{\mu}^{n-1}
\end{array} \right],
\]
and
\begin{eqnarray*}
\Sigma^{n,n}(x_*^{n},x_*^{n}) &=& k^{n,n}_{u,u}(x_*^{n},x_*^{n}) - \bm{q}^T\bm{K}^{-1}\bm{q} + \\
&&\bm{q}^T\bm{K}^{-1} \left[\begin{array}{cc}
0 & 0  \\ 
0 & \bm{\Sigma}^{n-1,n-1} 
\end{array} \right]\bm{K}^{-1}\bm{q}.
\end{eqnarray*}
Now, one can use the resulting posterior distribution (\ref{LinearMultistepMethods_posterior}) to obtain the artificially generated data $\{\bm{x}^{n},\bm{u}^{n}\}$ for the next time step with
\begin{equation}\label{eq:LinearMultistepMethods_artificial_data}
\bm{u}^{n} \sim \mathcal{N}\left(\bm{\mu}^{n}, \bm{\Sigma}^{n,n} \right).
\end{equation}
Here, $\bm{\mu}^{n} = \mu^{n}(\bm{x}^{n})$ and $\bm{\Sigma}^{n,n} = \Sigma^{n,n}(\bm{x}^{n},\bm{x}^{n})$.

\subsection{Example: Burgers' equation (Backward Euler)}\label{sec:Burgers}
\begin{figure}
\centering
\includegraphics[width=\textwidth]{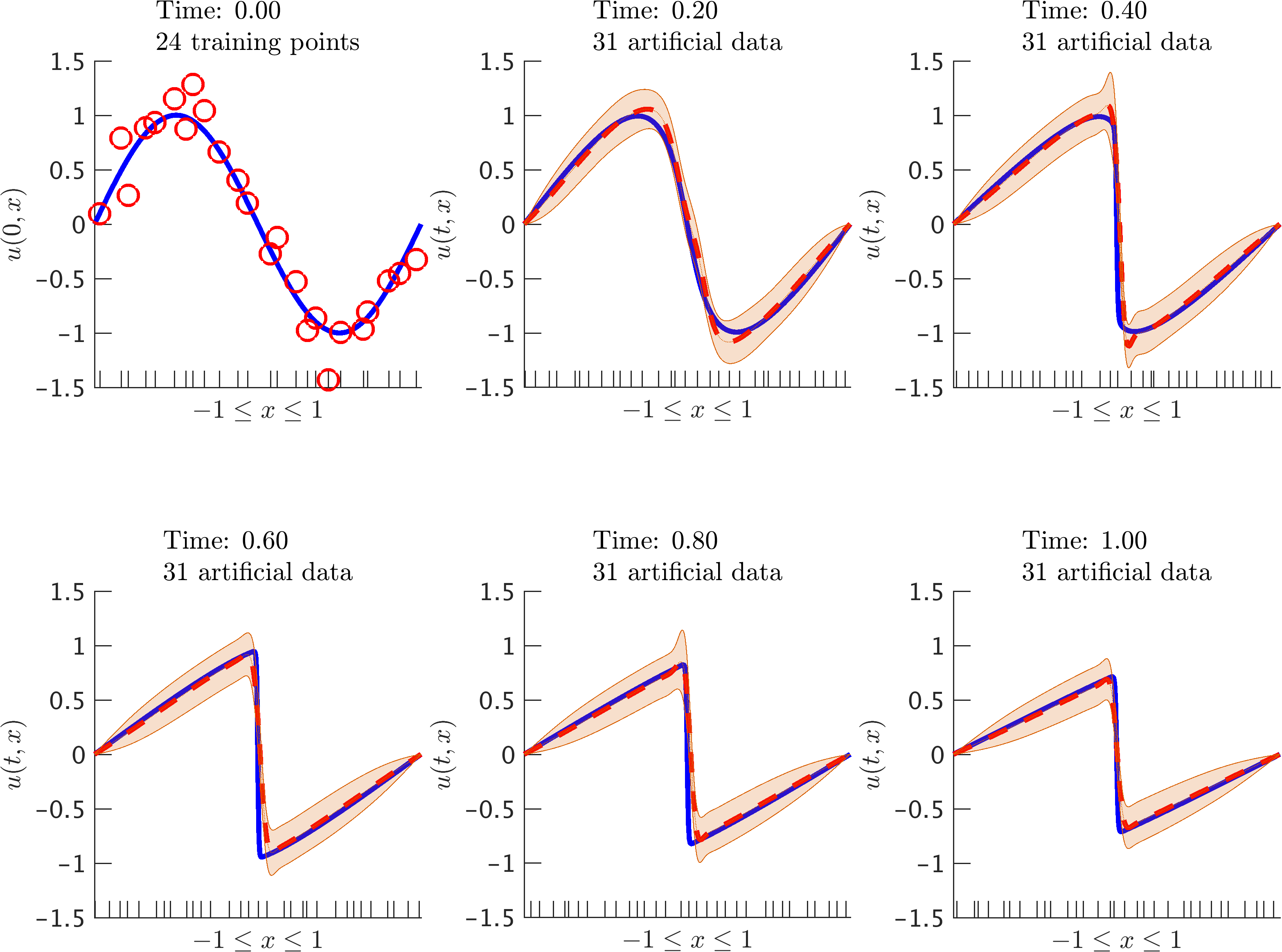}
\caption{\emph{Burgers' equation:} Initial data along with the posterior distribution of the solution at different time snapshots. The blue solid line represents the true data generating solution, while the dashed red line depicts the posterior mean. The shaded orange region illustrates the two standard deviations band around the mean. We are employing the backward Euler scheme with time step size $\Delta t = 0.01$. At each time step we generate $31$ artificial data points randomly located in the interval $[-1,1]$ according to a uniform distribution. These locations are highlighted by the ticks along the horizontal axis. Here, we set $\nu=0.01/\pi$ -- a value leading to the development of a non singular thin internal layer at $x=0$ that is notoriously hard to resolve by classical numerical methods \cite{basdevant1986spectral}. {\it (Code: \url{http://bit.ly/2mnUiKT}, Movie: \url{http://bit.ly/2m1sKHw})}}\label{fig:Burgers}
\end{figure}

Burgers' equation is a fundamental partial differential equation arising in various areas of applied mathematics, including fluid mechanics, nonlinear acoustics, gas dynamics, and traffic flow \cite{basdevant1986spectral}. In one space dimension the equation reads as
\begin{equation}\label{eq:Burgers}
u_t + u u_x = \nu u_{xx}, 
\end{equation}
along with Dirichlet boundary conditions $u(t,-1)=u(t,1)=0$, where $u(t,x)$ denotes the unknown solution and $\nu$ is a viscosity parameter. Let us assume that all we observe are noisy measurements $\{\bm{x}^0, \bm{u}^0\}$ of the \emph{black-box} initial function $u(0,x) = -\sin(\pi x)$. Given such measurements, we would like to solve the Burgers' equation (\ref{eq:Burgers}) while propagating through time the uncertainty associate with the noisy initial data (see figure \ref{fig:Burgers}).
This example is important because it involves solving a non-linear partial differential equation. To illustrate how one can encode the structure of the physical laws expressed by Burgers' equation in a \emph{numerical Gaussian process} let us apply the backward Euler scheme to equation (\ref{eq:Burgers}). This can be written as
\begin{equation}\label{eq:Burgers_Backward_Euler}
u^{n} = u^{n-1} - \Delta t u^{n} \frac{d}{d x}u^{n} + \nu \Delta t \frac{d^2}{d x^2} u^n.
\end{equation}
We would like to place a Gaussian process prior on $u^n$. However, the nonlinear term $u^{n} \frac{d}{d x}u^{n}$ is causing problems simply because the product of two Gaussian processes is no longer Gaussian. Hence, we will approximate the nonlinear term with $\mu^{n-1}\frac{d}{d x} u^{n}$, where $\mu^{n-1}$ is the posterior mean of the previous time step. Therefore, the backward Euler scheme (\ref{eq:Burgers_Backward_Euler}) can be approximated by
\begin{eqnarray}\label{eq:Burgers_approximate_backward_Euler}
u^{n} = u^{n-1} - \Delta t \mu^{n-1} \frac{d}{d x}u^{n} + \nu \Delta t \frac{d^2}{d x^2} u^n.
\end{eqnarray}
Rearranging the terms, we obtain
\begin{eqnarray}\label{eq:Burgers_approximate_backward_Euler_rearranged}
u^{n} + \Delta t \mu^{n-1} \frac{d}{d x}u^{n} - \nu \Delta t \frac{d^2}{d x^2} u^n =  u^{n-1}.
\end{eqnarray}

\subsubsection{Numerical Gaussian Process}
Let us make the prior assumption that
\begin{eqnarray}\label{eq:Burgers_prior}
u^{n}(x) \sim \mathcal{GP}(0, k(x,x';\theta)),
\end{eqnarray}
is a Gaussian process with a neural network \cite{Rasmussen06gaussianprocesses} covariance function
\begin{eqnarray}\label{eq:Burgers_neural_network}
k(x,x';\theta) =  \frac{2}{\pi}\sin^{-1}\left(\frac{2(\sigma_0^2 + \sigma^2 x x')}{\sqrt[]{(1+2\left(\sigma_0^2 + \sigma^2 x^2)\right)(1+2\left(\sigma_0^2 + \sigma^2 x'^2)\right)}}\right),
\end{eqnarray}
where $\theta = \left(\sigma^{2}_0,\sigma^{2}\right)$ denotes the hyper-parameters. Here we have chosen a non-stationary prior motivated by the fact that the solution to the Burgers' equation can develop discontinuities for small values of the viscosity parameter $\nu$. This enables us to obtain the following \emph{Numerical Gaussian Process}
\begin{equation*}
\left[\begin{array}{c}
u^n\\
u^{n-1}
\end{array}\right] \sim \mathcal{GP}\left(0,\left[\begin{array}{cc}
k^{n,n}_{u,u} & k^{n,n-1}_{u,u}\\
 & k^{n-1,n-1}_{u,u}
\end{array}\right]\right),
\end{equation*}
with covariance functions $k^{n,n}_{u,u}$, $k^{n,n-1}_{u,u}$, and $k^{n-1,n-1}_{u,u}$ given in section \ref{Appendix:Burgers} of the appendix. Training, prediction, and propagating the uncertainty associated with the noisy initial observations can be performed as in sections \ref{sec:Traning}, \ref{sec:Prediction}, and \ref{sec:Uncertainty}, respectively. Figure \ref{fig:Burgers} depicts the noisy initial data along with the posterior distribution of the solution to the Burgers' equation (\ref{eq:Burgers}) at different time snapshots. It is remarkable that the proposed methodology can effectively propagate an infinite collection of correlated Gaussian random variables (i.e., a Gaussian process) through the complex nonlinear dynamics of the Burgers' equation.

\subsubsection{Numerical Study}\label{sec:Burgers_numerical_study}
\begin{figure}
\centering
\includegraphics[width=\textwidth]{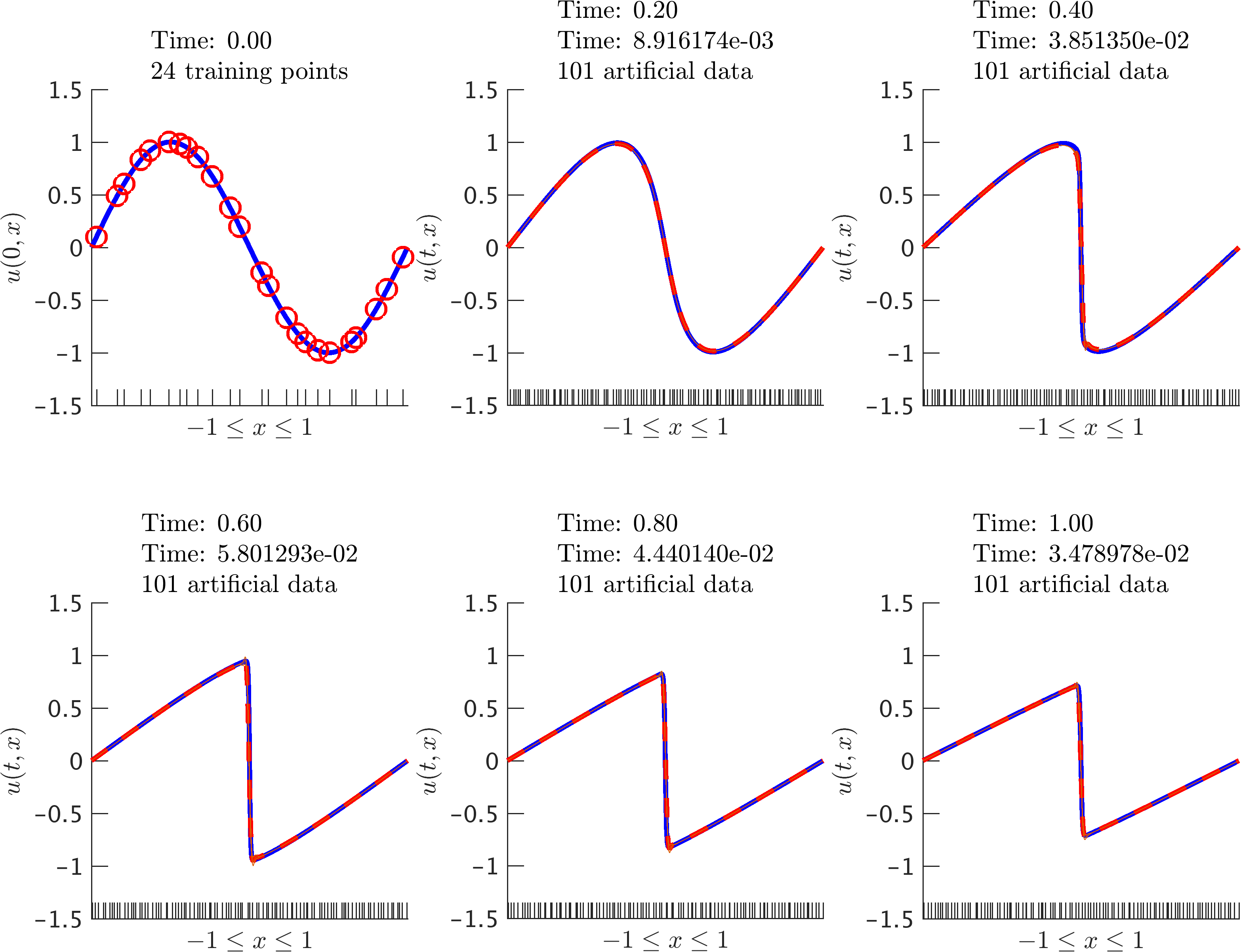}
\caption{\emph{Burgers' equation:} Initial data along with the posterior distribution of the solution at different time snapshots. The blue solid line represents the true data generating solution, while the dashed red line depicts the posterior mean. The shaded orange region illustrates the two standard deviations band around the mean. We are employing the backward Euler scheme with time step size $\Delta t = 0.01$. At each time step we generate $101$ artificial data points randomly located in the interval $[-1,1]$ according to a uniform distribution. These locations are highlighted by the ticks along the horizontal axis. Here, we set $\nu=0.01/\pi$ -- a value leading to the development of a non singular thin internal layer at $x=0$ that is notoriously hard to resolve by classical numerical methods \cite{basdevant1986spectral}. We are reporting the relative $\mathcal{L}^2$-error between the posterior mean and the true solution. {\it (Code: \url{http://bit.ly/2mDKCwb}, Movie: \url{http://bit.ly/2mDOPA5})}}\label{fig:Burgers_noiseless}
\end{figure}

\begin{figure}
\centering
\includegraphics[width=0.5\linewidth]{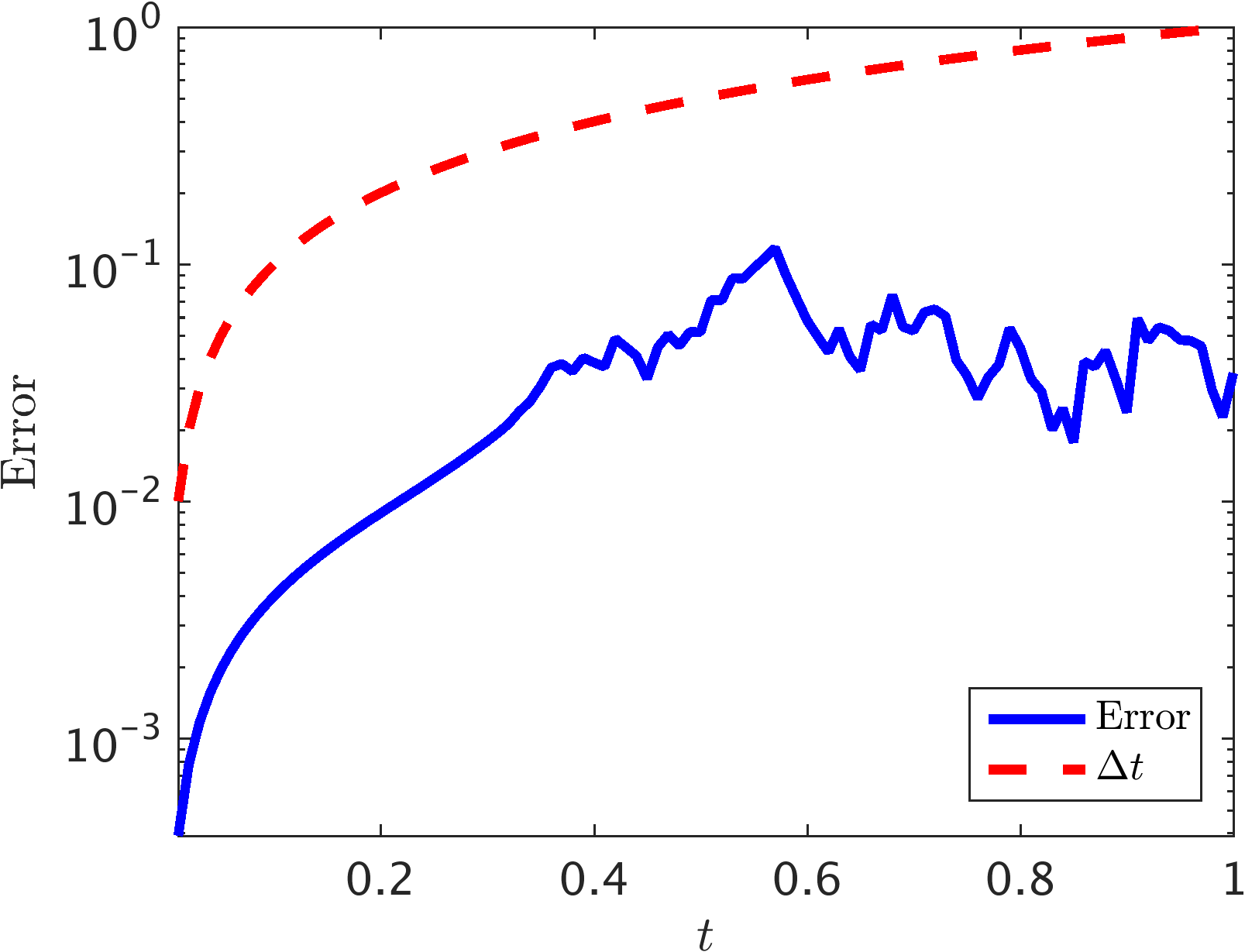}\\
\caption{\emph{Burgers' equation:} Time evolution of the relative spatial $\mathcal{L}^2$-error up to the final integration time $T=1.0$. We are using the backward Euler scheme with a time step-size of $\Delta t = 0.01$, and the red dashed line illustrates the optimal first-order convergence rate. {\it (Code: \url{http://bit.ly/2mDY6It})}}\label{fig:Burgers_error_versus_time}
\vspace*{0.5cm}
\minipage{0.48\textwidth}
  \includegraphics[width=\linewidth]{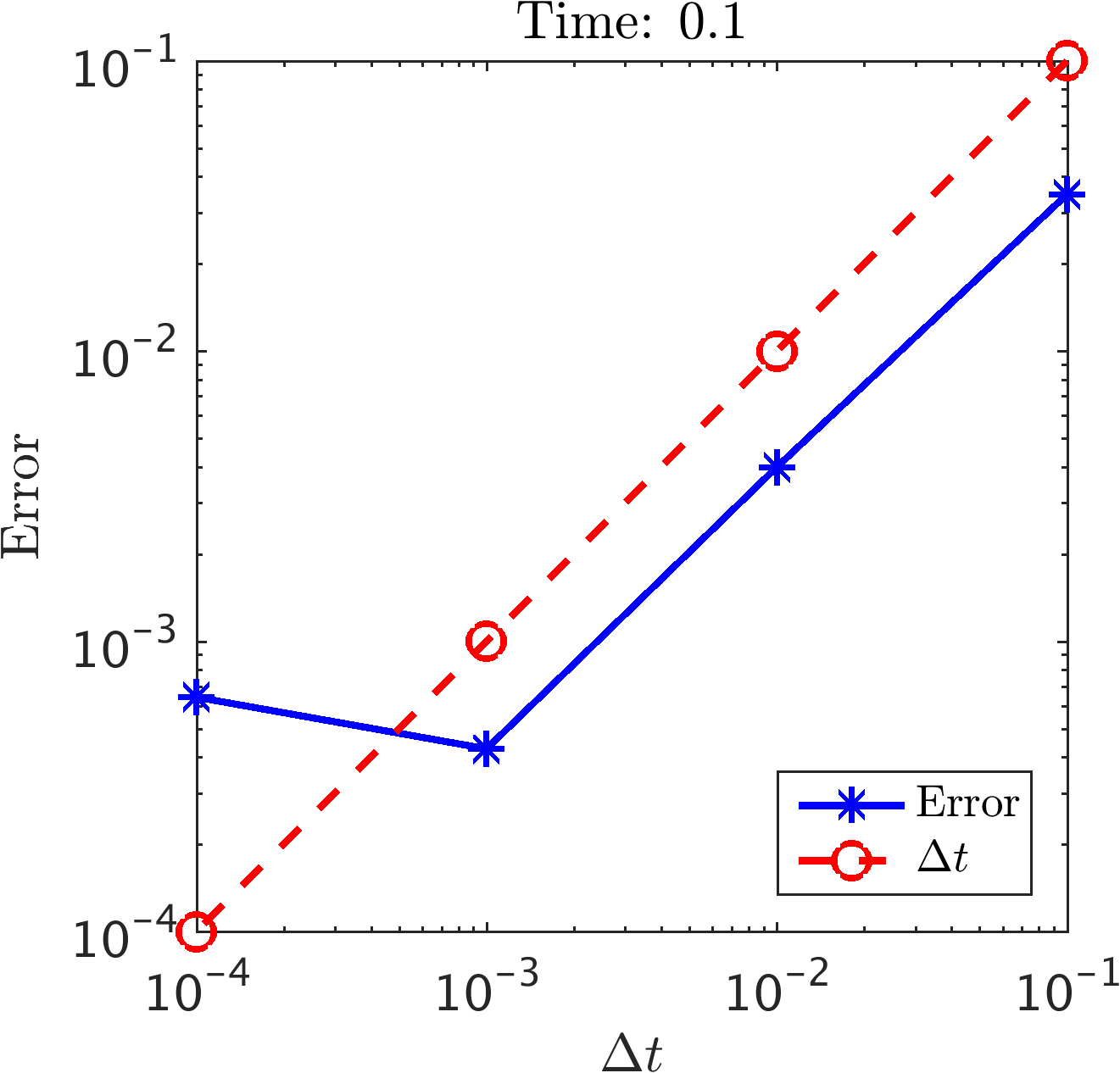}
  \caption{\emph{Burgers' equation:} Relative spatial $\mathcal{L}^2$-error versus step-size for the backward Euler scheme at time  $T=0.1$.  The number of  noiseless initial and artificially generated data is set to be equal to $50$. {\it (Code: \url{http://bit.ly/2mDY6It})}}\label{fig:Burgers_time_error}
\endminipage\hfill
\minipage{0.48\textwidth}
  \includegraphics[width=\linewidth]{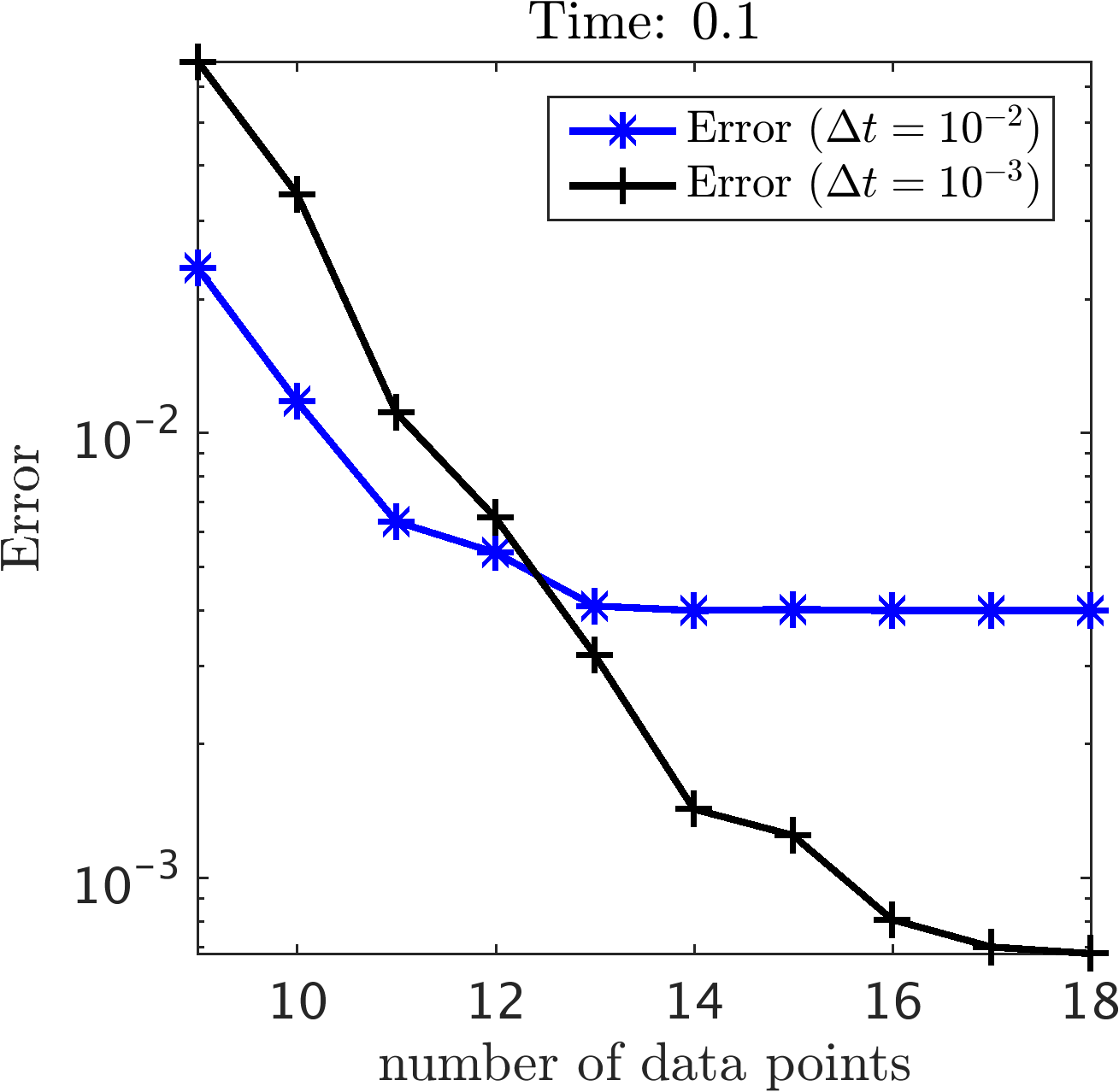}
  \caption{\emph{Burgers' equation:} Relative spatial $\mathcal{L}^2$-error versus the number of noiseless initial as well as artificial data points used for the backward Euler scheme with time step-sizes of $\Delta t = 10^{-2}$ and $\Delta t = 10^{-3}$. {\it (Code: \url{http://bit.ly/2mDY6It})}}\label{fig:Burgers_space_error}
\endminipage
\end{figure}

It must be re-emphasized that \emph{numerical Gaussian processes}, by construction, are designed to deal with cases where: (1) all we observe is noisy data on \emph{black-box} initial conditions, and (2) we are interested in \emph{quantifying the uncertainty} associated with such noisy data in our solutions to time-dependent partial differential equations. In fact, we recommend resorting to other alternative classical numerical methods such as Finite Differences, Finite Elements, and Spectral methods in cases where: (1) the initial function is \emph{not} a black-box function and we have access to \emph{noiseless} data, or (2) we are \emph{not} interested in quantifying the uncertainty in our solutions. However, in order to be able to perform a systematic numerical study of the proposed methodology and despite the fact that this defeats the whole purpose of the current work, sometimes we will operate under the assumption that we have access to \emph{noiseless} initial data. For instance, concerning the Burgers' equation, if we had access to such noiseless data, we would obtain results similar to the ones reported in figure \ref{fig:Burgers_noiseless}.
Moreover, in order to make sure that the \emph{numerical Gaussian process} resulting from the backward Euler scheme (\ref{eq:Burgers_approximate_backward_Euler_rearranged}) applied to the Burgers' equation is indeed first-order accurate in time, we perform the numerical experiments reported in figures \ref{fig:Burgers_error_versus_time} and \ref{fig:Burgers_time_error}.
Specifically, in figure \ref{fig:Burgers_error_versus_time} we report the time-evolution 
of the relative spatial $\mathcal{L}^{2}$ error until the final integration time $T=1.0$. 
We observe that the error indeed grows as $\mathcal{O}(\Delta{t})$, and its resulting behavior reveals both the shock development region as well as the energy dissipation due to diffusion at later times. Moreover, in figure \ref{fig:Burgers_time_error} we fix the final integration time to $T=0.1$ and the number of initial and artificial data to 50, and vary the time-step size $\Delta{t}$ from $10^{-1}$ to $10^{-4}$.  As expected, we recover the first-order convergence properties of the backward Euler scheme, except for a saturation  region arising when we further reduce the time-step size below approximately $10^{-3}$. This behavior is not a result of the time stepping scheme but is attributed to the underlying Gaussian process regression and the finite number of spatial data points used for training and prediction. 
%Indeed, if we double the number of training points this saturation occurs at lower time-step sizes, but here we have chosen to deliberately demonstrate this interplay between spatial and temporal accuracy.  
To  investigate the accuracy of the posterior mean in predicting the solution as the number of training points is increased, we perform the numerical experiment reported in figure \ref{fig:Burgers_space_error}. Here we have considered two cases for which we fix the time step size to $\Delta{t}=10^{-2}$ and $\Delta{t}=10^{-3}$, respectively, and increase the number of initial as well as artificial data points. A similar accuracy saturation is also observed here as the number of training points is increased. In this case, this is attributed to the error accumulated due to time-stepping with the relatively large time step sizes for the first-order accurate Euler scheme. If we further keep decreasing the time-step, this saturation behavior will occur for higher numbers of total training points. The key point here is that although Gaussian processes can yield satisfactory accuracy, they, by construction, cannot force the approximation error down to machine precision. This is due to the fact that Gaussian processes are suitable for solving regression problems. This is exactly the reason why we recommend other alternative classical numerical methods for solving partial differential equations in cases where one has access to noiseless data. In such cases, it is desirable to use numerical schemes that are capable of performing exact interpolation on the data, rather than just a mere regression.

\subsection{Example: Wave Equation (Trapezoidal Rule)}
\begin{figure}
\centering
\includegraphics[width=\textwidth]{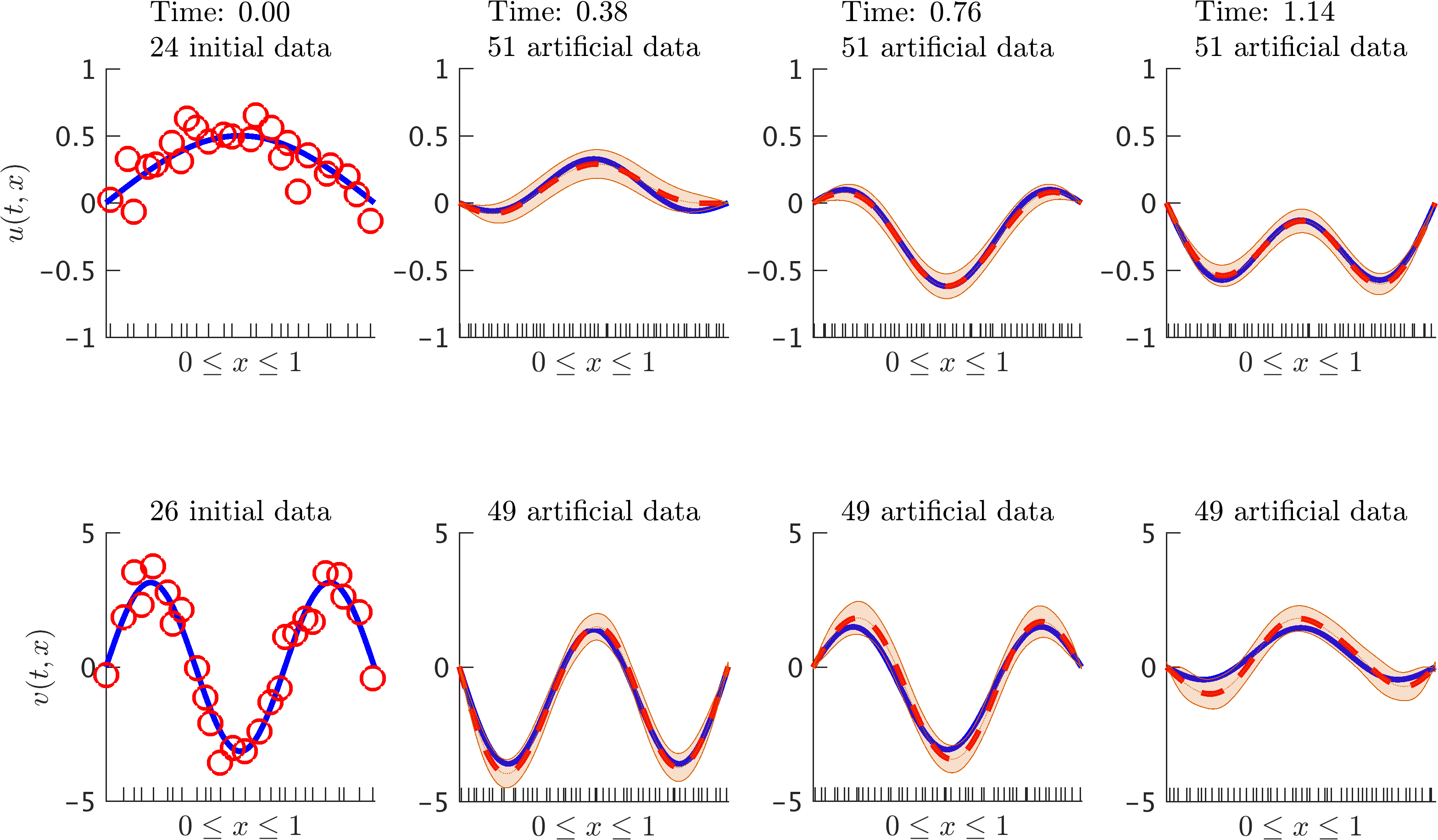}
\caption{\emph{Wave equation:} Initial data along with the posterior distribution of the solution at different time snapshots. Here, $v(t,x) = u_t(t,x)$. The blue solid line represents the true data generating solution, while the dashed red line depicts the posterior mean. The shaded orange region illustrates the two standard deviations band around the mean. At each time step we generate $51$ artificial data points for $u$ and 49 for $v$, all randomly located in the interval $[0,1]$ according to a uniform distribution. These locations are highlighted by the ticks along the horizontal axis. We are employing the trapezoidal scheme with time step size $\Delta t = 0.01$. {\it (Code: \url{http://bit.ly/2m3mfnA}, Movie: \url{http://bit.ly/2mpfhNi})}}\label{fig:Wave}
\end{figure}

The wave equation is an important second-order linear partial differential equation for the description of wave propagation phenomena, including sound waves, light waves, and water waves. It arises in many scientific fields such as acoustics, electromagnetics, and fluid dynamics. In one space dimension the wave equation reads as
\begin{equation}\label{eq:Wave}
u_{tt} = u_{xx}.
\end{equation}
The function $u(t,x) = \frac12 \sin(\pi x) \cos(\pi t) + \frac13 \sin(3 \pi x )\sin(3 \pi t)$ solves this equation and satisfies the following initial and homogeneous Dirichlet boundary conditions
\begin{eqnarray}\label{eq:Wave_initial_boundary}
&&u(0, x) = u^0(x) := \frac12 \sin(\pi x),\nonumber\\
&&u_t(0, x) =  v^0(x) := \pi \sin(3 \pi x),\nonumber\\
&&u(t,0) = u(t,1) = 0.
\end{eqnarray}
Now, let us assume that all we observe are noisy measurements $\{\bm{x}^0_u, \bm{u}^0\}$ and $\{\bm{x}^0_v, \bm{v}^0\}$ of the \emph{black-box} initial functions $u^0$ and $v^0$, respectively. Given this data, we are interested in solving the wave equation (\ref{eq:Wave}) and quantifying the uncertainty in our solution associated with the noisy initial data (see figure \ref{fig:Wave}).
To proceed, let us define $v := u_t$ and rewrite the wave equation as a system of equations given by
\begin{equation}\label{eq:Wave_system}
\left\{\begin{array}{l}
u_t = v,\\
v_t = u_{xx}.\\
\end{array}\right.
\end{equation}
This example is important because it involves solving a \emph{system} of partial differential equations. One could rewrite the system of equations (\ref{eq:Wave_system}) in matrix-vector notations and obtain
\[
\frac{\partial}{\partial t} \left[\begin{array}{c}
u\\
v
\end{array}\right] = \mathcal{L}_x \left[\begin{array}{c}
u\\
v
\end{array}\right],
\]
which takes the form of (\ref{eq:Linear}) with
\[
\mathcal{L}_x = \left[\begin{array}{cc}
0 & I\\
\frac{\partial^2}{\partial x^2} & 0
\end{array}\right].
\]
This form is now amenable to the previous analysis provided for general linear multi-step methods. However, for pedagogical purposes, let us walk slowly through the trapezoidal rule and apply it to the system of equations (\ref{eq:Wave_system}). This can be written as
\begin{eqnarray}\label{eq:Wave_trapezoidal}
u^{n} &=& u^{n-1} + \frac12 \Delta t v^{n-1} + \frac12 \Delta t v^{n},\\
v^{n} &=& v^{n-1} + \frac12 \Delta t \frac{d^2}{d x^2}u^{n-1} + \frac12 \Delta t \frac{d^2}{d x^2}u^{n}.\nonumber
\end{eqnarray}
Rearranging the terms yields
\begin{eqnarray*}
u^{n} - \frac12 \Delta t v^{n} &=& u^{n-1} + \frac12 \Delta t v^{n-1},\\
v^{n} - \frac12 \Delta t \frac{d^2}{d x^2}u^{n} &=& v^{n-1} + \frac12 \Delta t \frac{d^2}{d x^2}u^{n-1}.\nonumber
\end{eqnarray*}
Now, let us define $u^{n-1/2}$ and $v^{n-1/2}$ to be given by
\begin{eqnarray}\label{eq:Wave_trapezoidal_half_way}
u^{n} - \frac12 \Delta t v^{n} &=:& u^{n-1/2} := u^{n-1} + \frac12 \Delta t v^{n-1},\\
v^{n} - \frac12 \Delta t \frac{d^2}{d x^2}u^{n} &=:& v^{n-1/2} := v^{n-1} + \frac12 \Delta t \frac{d^2}{d x^2}u^{n-1}.\nonumber
\end{eqnarray}
As outlined in section \ref{sec:LinearMultistepMethods} this is a key step in the proposed methodology as it hints at the proper location to place the Gaussian process prior. Shifting the terms involved in the above equations by $-1/2$ and $+1/2$ we obtain
\begin{eqnarray}\label{eq:Wave_trapezoidal_shifted_left}
u^{n-1/2} - \frac12 \Delta t v^{n-1/2} &=& u^{n-1},\\
v^{n-1/2} - \frac12 \Delta t \frac{d^2}{d x^2}u^{n-1/2} &=& v^{n-1},\nonumber
\end{eqnarray}
and
\begin{eqnarray}\label{eq:Wave_trapezoidal_shifted_right}
u^{n} &=& u^{n-1/2} + \frac12 \Delta t v^{n-1/2},\\
v^{n} &=& v^{n-1/2} + \frac12 \Delta t \frac{d^2}{d x^2}u^{n-1/2},\nonumber
\end{eqnarray}
respectively. Now we can proceed with encoding the structure of the wave equation into a \emph{numerical Gaussian process} prior for performing Bayesian machine learning of the solution $\{u(t,x),v(t,x)\}$ at any $t>0$.

\subsubsection{Numerical Gaussian Process}
Let us make the prior assumption that
\begin{eqnarray}\label{eq:Wave_trapezoidal_prior}
u^{n-1/2}(x) &\sim & \mathcal{GP}(0, k_u(x,x';\theta_u)),\\
v^{n-1/2}(x) &\sim & \mathcal{GP}(0, k_v(x,x';\theta_v)),\nonumber
\end{eqnarray}
are two independent Gaussian processes with squared exponential \cite{Rasmussen06gaussianprocesses} covariance functions
\begin{eqnarray}\label{eq:Wave_squared_exponential}
k_u(x,x';\theta_u) =  \gamma_{u}^2 \exp\left(-\frac12 w_u(x - x')^2\right),\\
k_v(x,x';\theta_v) =  \gamma_{v}^2 \exp\left(-\frac12 w_v(x - x')^2\right),\nonumber
\end{eqnarray}
where $\theta_u = \left(\gamma_{u}^{2},w_u\right)$ and $\theta_v = \left(\gamma_{v}^{2},w_v\right)$. From a theoretical point of view, each covariance function gives rise to a Reproducing Kernel Hilbert space \cite{aronszajn1950theory, saitoh1988theory, berlinet2011reproducing} that defines a class of functions that can be represented by this kernel. In particular, the squared exponential covariance function chosen above implies smooth approximations. More complex function classes can be accommodated by appropriately choosing kernels (see e.g., equation (\ref{eq:Burgers_neural_network})).  This enables us to obtain the following \emph{numerical Gaussian process}
\begin{eqnarray*}
\left[\begin{array}{c}
u^n\\
v^{n}\\
u^{n-1}\\
v^{n-1}\\
\end{array}\right] \sim \mathcal{GP}\left(0, \left[\begin{array}{cccc}
k^{n,n}_{u,u} & k^{n,n}_{u,v} & k^{n,n-1}_{u,u} & k^{n,n-1}_{u,v}\\
& k^{n,n}_{v,v} & k^{n,n-1}_{v,u} & k^{n,n-1}_{v,v}\\
 & & k^{n-1,n-1}_{u,u} & k^{n-1,n-1}_{u,v}\\
 &  &  & k^{n-1,n-1}_{v,v}\\
\end{array}\right]\right),
\end{eqnarray*}
which captures the entire structure of the trapezoidal rule (\ref{eq:Wave_trapezoidal}), applied to the wave equation (\ref{eq:Wave}), in its covariance functions given in section \ref{Appendix:Wave} of the appendix. Training, prediction, and propagating the uncertainty associated with the noisy initial observations can be performed as in section \ref{Appendix:Wave} of the appendix. Figure \ref{fig:Wave} depicts the noisy initial data along with the posterior distribution (\ref{eq:Wave_posterior}) of the solution to the wave equation (\ref{eq:Wave}) at different time snapshots.

\subsubsection{Numerical Study}
\begin{figure}
\centering
\includegraphics[width=\textwidth]{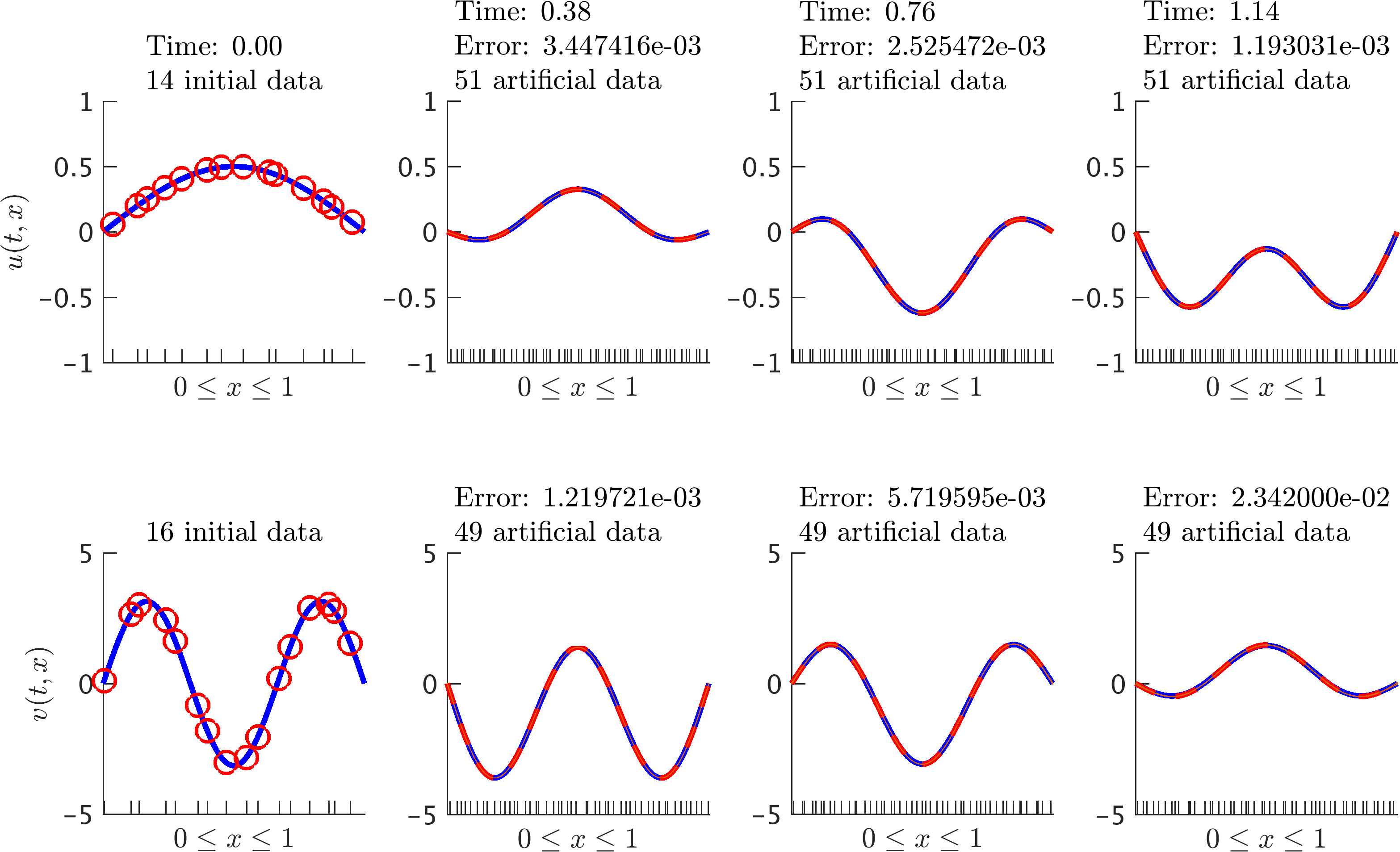}
\caption{\emph{Wave equation:} Initial data along with the posterior distribution of the solution at different time snapshots. Here, $v(t,x) = u_t(t,x)$. The blue solid line represents the true data generating solution, while the dashed red line depicts the posterior mean. The shaded orange region illustrates the two standard deviations band around the mean. At each time step we generate $51$ artificial data points for $u$ and 49 for $v$, all randomly located in the interval $[0,1]$ according to a uniform distribution. These locations are highlighted by the ticks along the horizontal axis. We are employing the trapezoidal scheme with time step size $\Delta t = 0.01$. We are reporting the relative $\mathcal{L}^2$-error between the posterior mean and the true solution. {\it (Code: \url{http://bit.ly/2m3mKhK}, Movie: \url{http://bit.ly/2mFalVg})}}\label{fig:Wave_noiseless}
\end{figure}

\begin{figure}
\centering
\includegraphics[width=0.5\linewidth]{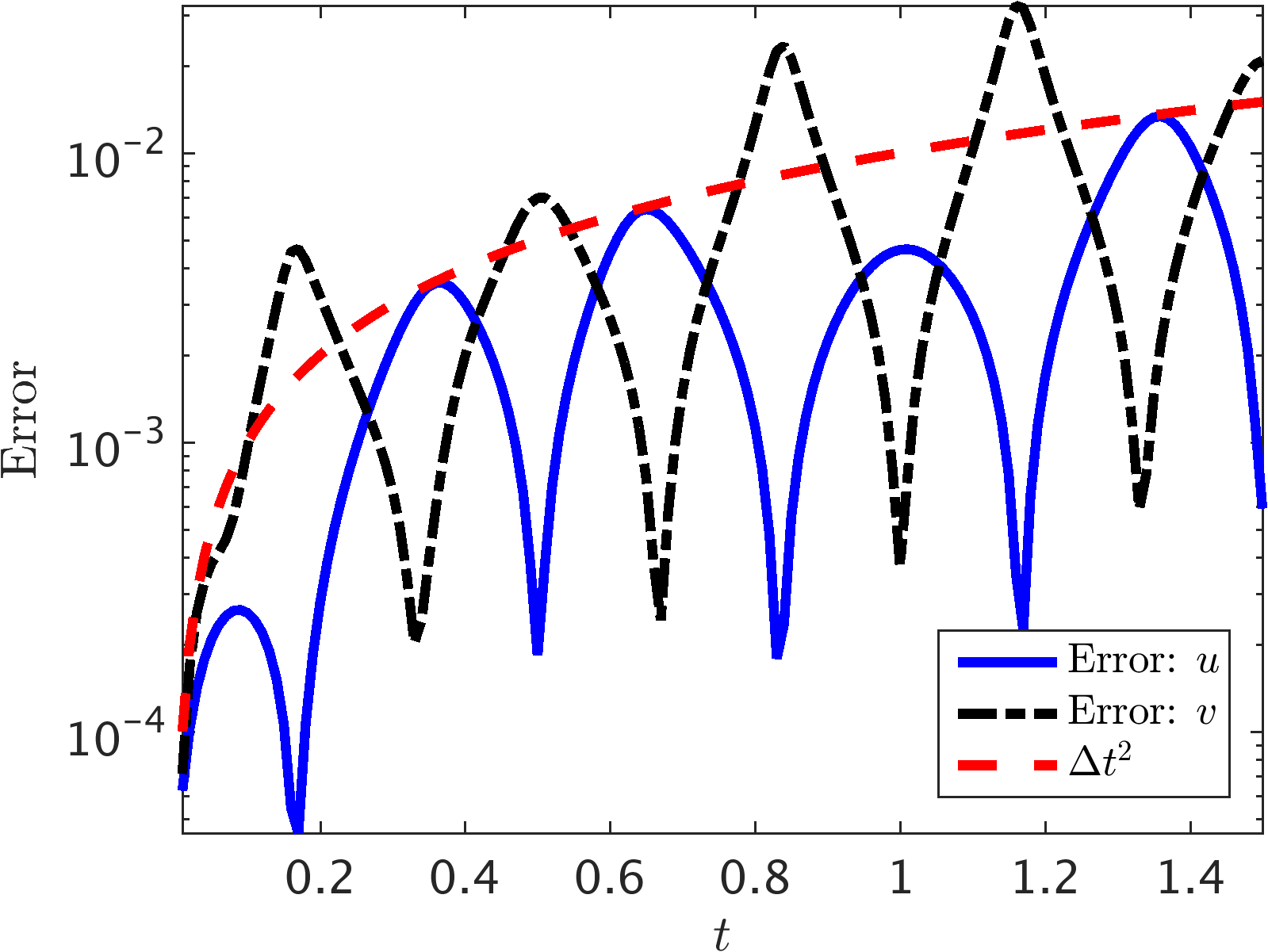}\\
\caption{\emph{Wave equation:} Time evolution of the relative spatial $\mathcal{L}^2$-error up to the final integration time $T=1.5$. The blue solid line corresponds to the $u$ component of the solution while the black dashed line corresponds to the function $v$. We are using the trapezoidal rule with a time step-size of $\Delta t = 0.01$, and the red dashed line illustrates the optimal second-order convergence rate. {\it (Code: \url{http://bit.ly/2niW6lW})}}\label{fig:Wave_error_versus_time}
\vspace*{0.5cm}
\minipage{0.48\textwidth}
  \includegraphics[width=\linewidth]{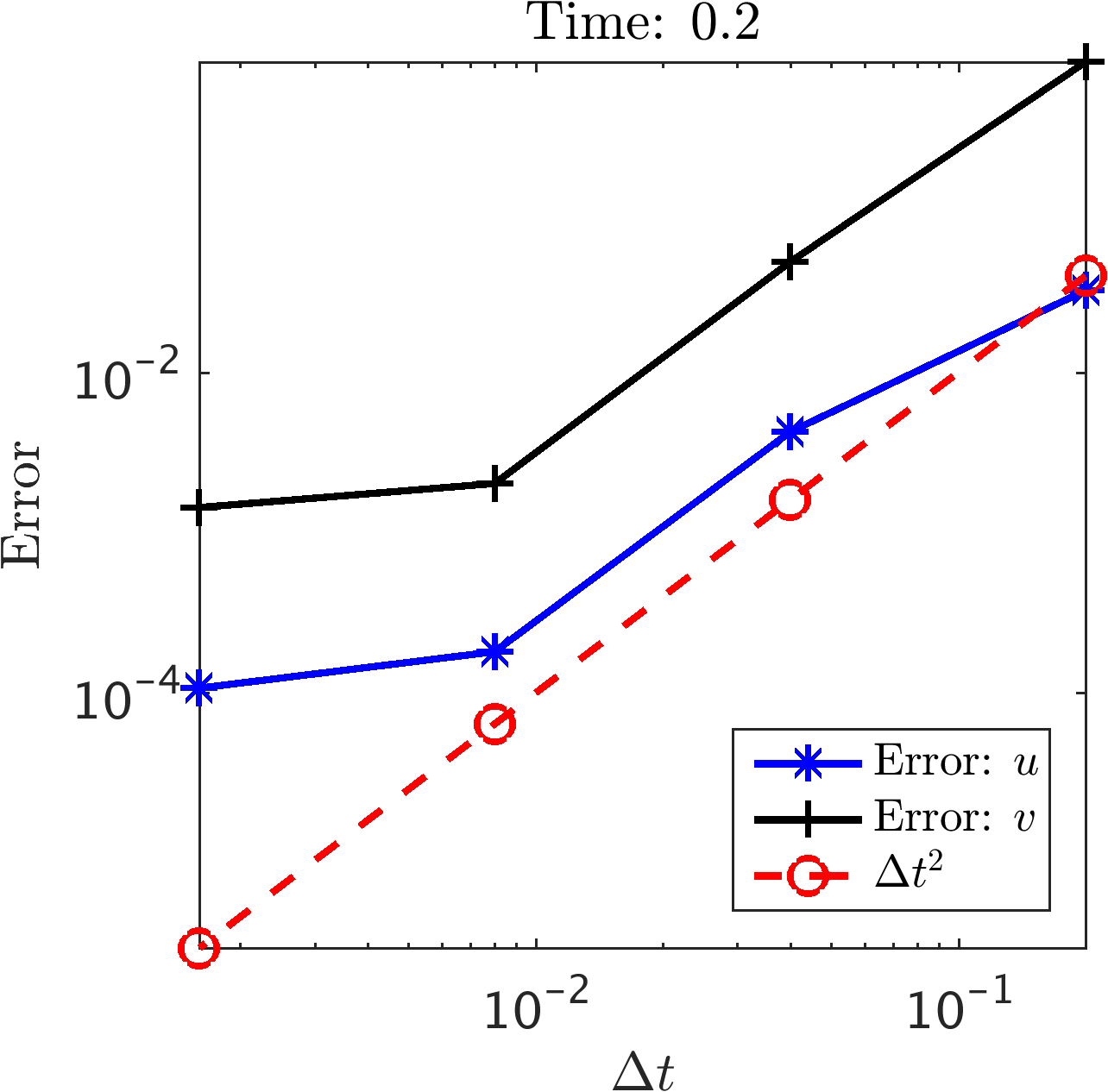}
  \caption{\emph{Wave equation:} Relative spatial $\mathcal{L}^2$-error versus step-size for the trapezoidal rule. Here, the number of noiseless initial data as well as the artificially generated data is set to be equal to $50$. We are running the time stepping scheme up until time $0.2$. {\it (Code: \url{http://bit.ly/2niW6lW})}}\label{fig:Wave_time_error}
\endminipage\hfill
\minipage{0.48\textwidth}
  \includegraphics[width=\linewidth]{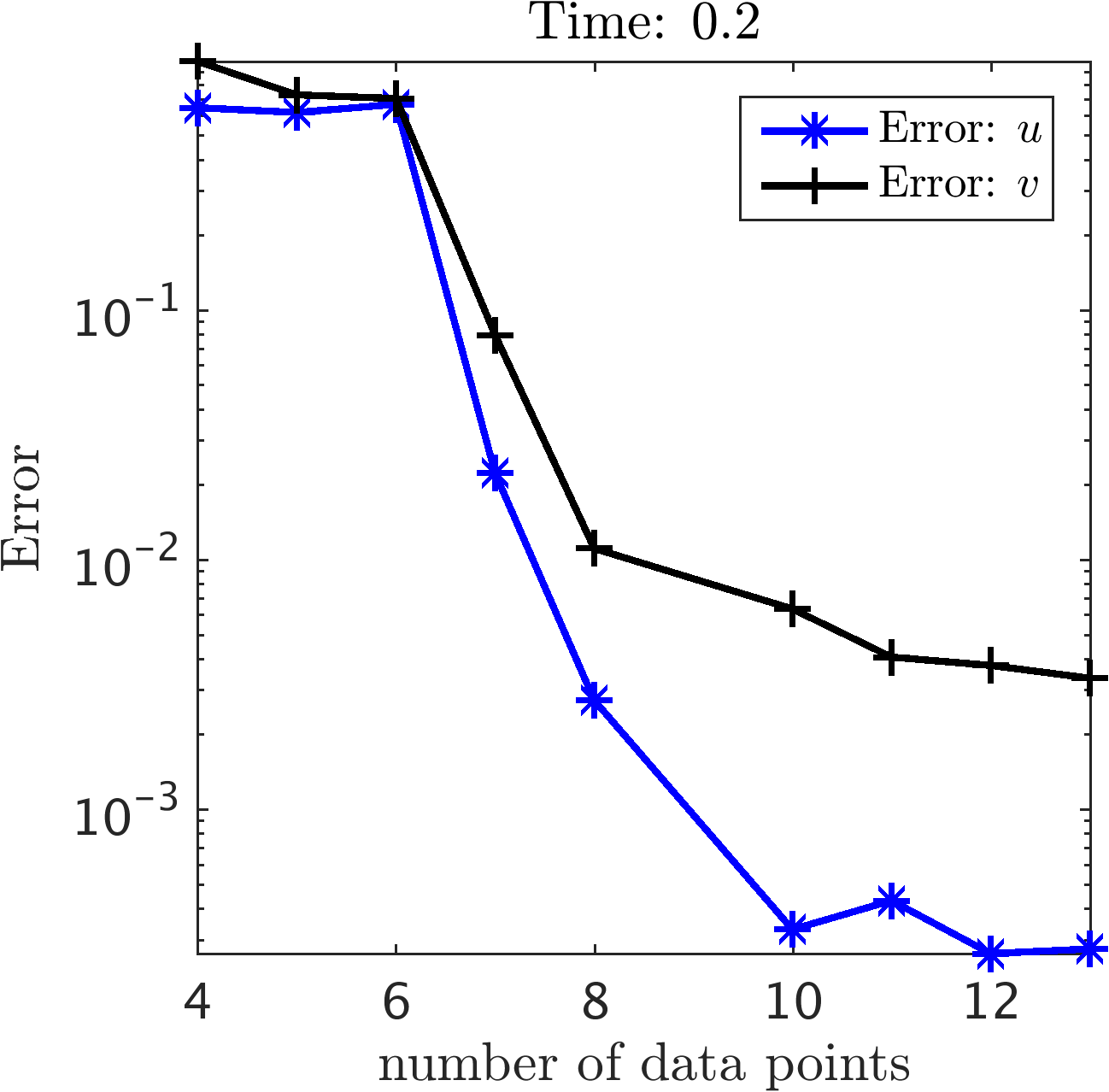}
  \caption{\emph{Wave equation:} Relative spatial $\mathcal{L}^2$-error versus the number of noiseless initial as well as artificial data points used for the trapezoidal rule. Here, the time step-size is set to be $\Delta t = 0.01$. We are running the time stepping scheme up until time $0.2$. {\it (Code: \url{http://bit.ly/2niW6lW})}}\label{fig:Wave_space_error}
\endminipage
\end{figure}

In the case where we have access to noiseless initial data we obtain the results depicted in figure \ref{fig:Wave_noiseless}.
Moreover, we perform a numerical study similar to the one reported in section \ref{sec:Burgers_numerical_study}. This is to verify that the \emph{numerical Gaussian process} resulting from the trapezoidal rule (\ref{eq:Wave_trapezoidal}) applied to the wave equation is indeed second-order accurate in time. In particular, the numerical experiment shown in figure \ref{fig:Wave_error_versus_time} illustrates the time evolution of the relative spatial $\mathcal{L}^{2}$ until the final integration time $T=1.5$. The second-order convergence of the algorithm is also demonstrated in figure \ref{fig:Wave_time_error} where we have fixed the number of noiseless initial and artificially generated data, while decreasing the time step size. We also investigate the convergence behavior of the algorithm for a fixed time-step $\Delta{t}=10^{-2}$ and as the number of training points is increased. The results are summarized in figure \ref{fig:Wave_space_error}. The analysis of both temporal and spatial convergence properties yield qualitatively similar conclusions to the ones reported in section \ref{sec:Burgers_numerical_study}. One thing worth mentioning here is that the error in $u$ is not always less than the error in $v$ (as seen in figures \ref{fig:Wave_noiseless} and \ref{fig:Wave_error_versus_time}). This just happens to be the case at time $T = 0.2$.

\section{Runge-Kutta Methods} 
Let us now focus on the general form of Runge-Kutta methods \cite{alexander1977diagonally} with $q$ stages applied to equation (\ref{eq:Linear}); i.e.,
\begin{eqnarray}\label{eq:RungeKutta}
u^{n+1} &=& u^{n} + \Delta t \sum_{i=1}^q b_i \mathcal{L}_x u^{n+\tau_i},\\
u^{n+\tau_i} &=& u^n + \Delta t \sum_{j=1}^q a_{ij} \mathcal{L}_x u^{n+\tau_j}, \ \ i=1,\ldots,q.\nonumber
\end{eqnarray}
Here, $u^{n+\tau_i}(x) = u(t^n + \tau_i \Delta t, x)$. This general form encapsulates both implicit and explicit time-stepping schemes, depending on the choice of the weights $\{a_{ij},b_i\}$. An important feature of the proposed methodology is that it is oblivious to the choice of these parameters, hence the implicit or explicit nature of the time-stepping scheme is ultimately irrelevant. This is in sharp contrast to classical numerical methods in which implicit time-integration is burdensome due to the need for repeatedly solving linear or nonlinear systems. Here, for a fixed number of stages $q$, the cost of performing implicit or explicit time-marching is identical. This is attributed to the fact that the structure of the time-stepping scheme is encoded in the \emph{numerical Gaussian process} prior, and the algorithm only involves solving a sequence of regression problems as outlined in section \ref{sec:Workflow}. This allows us to enjoy the favorable stability properties of fully implicit schemes at no extra cost, and thus perform long-time integration using very large time-steps. Equations (\ref{eq:RungeKutta}) can be equivalently written as
\begin{eqnarray}\label{eq:RungeKutta_operator}
u^{n+1} - \Delta t \sum_{i=1}^q b_i \mathcal{L}_x u^{n+\tau_i} &=& u^n =: u^n_{q+1},\\
u^{n+\tau_i} - \Delta t \sum_{j=1}^q a_{ij} \mathcal{L}_x u^{n+\tau_j} &=& u^n =: u^n_i, \ \ i=1,\ldots,q.\nonumber
\end{eqnarray}
Let us make the prior assumption that
\begin{eqnarray}\label{eq:RungeKutta_prior_assumption}
u^{n+1}(x) &\sim & \mathcal{GP}(0,k^{n+1,n+1}_{u,u}(x,x';\theta_{n+1})),\\
u^{n+\tau_i}(x) &\sim & \mathcal{GP}(0,k^{n+\tau_i,n+\tau_i}_{u,u}(x,x';\theta_{n+\tau_i})), \ \ \ i=1,\ldots,q,\nonumber
\end{eqnarray}
are $q+1$ mutually independent Gaussian processes. Therefore, we can write the joint distribution of $u^{n+1}, u^{n+\tau_q}, \ldots, u^{n+\tau_1}, u^{n}_{q+1}, \ldots, u^{n}_1$ which will capture the entire structure of the Runge-Kutta methods in the resulting \emph{numerical Gaussian process}. However, rather than getting bogged down into heavy notation, and without sacrificing any generality, we will present the main ideas through the lens of an example. 

\subsection{Example: Advection Equation (Gauss-Legendre Method)}\label{sec:Advection}
\begin{figure}
\centering
\includegraphics[width=\textwidth]{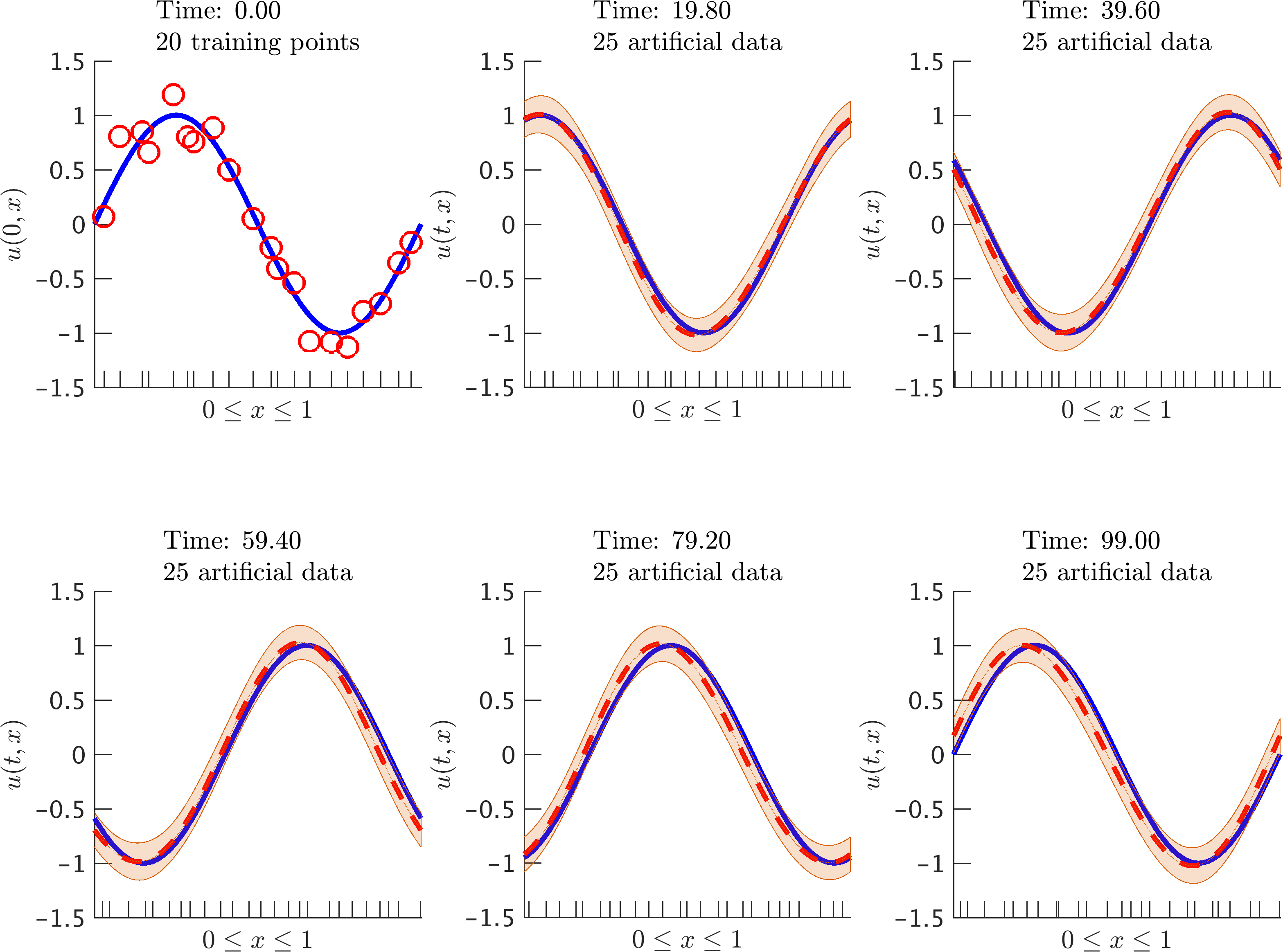}
\caption{\emph{Advection equation:} Initial data along with the posterior distribution of the solution at different time snapshots. The blue solid line represents the true data generating solution, while the dashed red line depicts the posterior mean. The shaded orange region illustrates the two standard deviations band around the mean. At each time step we generate $25$ artificial data points  randomly located in the interval $[0,1]$ according to a uniform distribution. These locations are highlighted by the ticks along the horizontal axis.  We are employing the Gauss-Legendre time-stepping quadrature rule with time step size $\Delta t = 0.1$. It is worth highlighting that we are running the time stepping scheme for a very long time and with a relatively large time step size. {\it (Code: \url{http://bit.ly/2m3JoXb}, Movie: \url{http://bit.ly/2mKHCP4})}}\label{fig:Advection}
\end{figure}

We have chosen this classical pedagogical example as a prototype benchmark problem for testing the limits of long-time integration. This example also highlights the implementation of periodic constraints at the domain boundaries (\ref{eq:Advection_initial_boundary}). The advection equation in one space dimension takes the form
\begin{eqnarray}\label{eq:Advection}
u_t =  -u_x.
\end{eqnarray}
The function $u(t,x) = \sin(2\pi(x-t))$ solves this equation and satisfies the following initial and periodic boundary conditions
\begin{eqnarray}\label{eq:Advection_initial_boundary}
u(0,x) &=& u^0(x) := \sin(2\pi x),\nonumber\\
u(t,0) &=& u(t,1).
\end{eqnarray}
However, let us assume that all we observe are noisy measurements $\{\bm{x}^0, \bm{u}^0\}$ of the \emph{black-box} initial function $u^0$. Given this data, we are interested in encoding the structure of the advection operator in a \emph{numerical Gaussian process} prior and use it to infer the solution $u(t,x)$ with quantified uncertainty for any $t>0$  (see figure \ref{fig:Advection}).
 Let us apply the Gauss-Legendre time-stepping quadrature \cite{iserles2009first} with two stages (thus fourth-order accurate) to the advection equation (\ref{eq:Advection}). Referring to equations (\ref{eq:RungeKutta_operator}), we obtain
\begin{eqnarray}\label{eq:GaussLegendre}
u^n_3 := u^n &=& u^{n+1} + b_1 \Delta t \frac{d}{d x}u^{n+\tau_1} + b_2 \Delta t  \frac{d}{d x}u^{n+\tau_2},\\
u^n_2 := u^n &=& u^{n+\tau_2} + a_{21} \Delta t \frac{d}{d x}u^{n+\tau_1} + a_{22} \Delta t  \frac{d}{d x}u^{n+\tau_2},\nonumber\\
u^n_1 := u^n &=& u^{n+\tau_1} + a_{11} \Delta t \frac{d}{d x}u^{n+\tau_1} + a_{12} \Delta t  \frac{d}{d x}u^{n+\tau_2}.\nonumber
\end{eqnarray}
Here, $\tau_1 = \frac{1}{2} - \frac{1}{6}\sqrt{3}$, $\tau_2 = \frac{1}{2} + \frac{1}{6}\sqrt{3}$, $b_1 = b_2 = \frac{1}{2}$, $a_{11} = a_{22} = \frac{1}{4}$, $a_{12} =  \frac{1}{4} - \frac{1}{6}\sqrt{3}$, and $a_{21} =  \frac{1}{4} + \frac{1}{6}\sqrt{3}$.

\subsubsection{Prior}
We make the prior assumption that
\begin{eqnarray}
u^{n+1}(x) &\sim & \mathcal{GP}(0, k^{n+1,n+1}_{u,u}(x,x';\theta_{n+1})),\\
u^{n+\tau_2}(x) &\sim & \mathcal{GP}(0, k^{n+\tau_2,n+\tau_2}_{u,u}(x,x';\theta_{n+\tau_2})),\nonumber\\
u^{n+\tau_1}(x) &\sim & \mathcal{GP}(0, k^{n+\tau_1,n+\tau_1}_{u,u}(x,x';\theta_{n+\tau_1})),\nonumber
\end{eqnarray}
are three independent Gaussian processes with squared exponential covariance functions similar to the kernels used in equations (\ref{eq:Wave_squared_exponential}). This assumption yields the following \emph{numerical Gaussian process}
\[
\resizebox{\textwidth}{!}{$
\left[\begin{array}{c}
u^{n+1} \\ 
u^{n+\tau_2} \\
u^{n+\tau_1} \\ 
u^n_{3} \\ 
u^n_{2} \\ 
u^n_{1}
\end{array} \right] \sim \mathcal{GP}\left(0, \left[\begin{array}{cccccc}
k^{n+1,n+1}_{u,u} & 0 & 0 & k^{n+1,n}_{u,3} & 0 & 0 \\ 
 & k^{n+\tau_2,n+\tau_2}_{u,u} & 0 & k^{n+\tau_2,n}_{u,3} &  k^{n+\tau_2,n}_{u,2} & k^{n+\tau_2,n}_{u,1} \\ 
 &  & k^{n+\tau_1,n+\tau_1}_{u,u} & k^{n+\tau_1,n}_{u,3} &  k^{n+\tau_1,n}_{u,2} & k^{n+\tau_1,n}_{u,1} \\ 
 &  &  & k^{n,n}_{3,3} &  k^{n,n}_{3,2} & k^{n,n}_{3,1} \\ 
 &  & & & k^{n,n}_{2,2} & k^{n,n}_{2,1} \\ 
 &  &   & & & k^{n,n}_{1,1}
\end{array} \right]\right),
$}
\]
where the covariance functions are given in section \ref{Appendix:Advection} of the appendix.

\subsubsection{Training}
The hyper-parameters $\theta_{n+1}$, $\theta_{n+\tau_2}$, and $\theta_{n+\tau_1}$ can be trained by minimizing the Negative Log Marginal Likelihood resulting from
\begin{equation}\label{eq:Advection_NLML}
\left[\begin{array}{c}
u^{n+1}(1) - u^{n+1}(0) \\ 
u^{n+\tau_2}(1) - u^{n+\tau_2}(0)\\
u^{n+\tau_1}(1) - u^{n+\tau_1}(0)\\
\bm{u}_3^{n} \\ 
\bm{u}_2^{n} \\
\bm{u}_1^{n}
\end{array} \right] \sim \mathcal{N}\left(0,\bm{K}\right).
\end{equation}
Here, $u^{n+1}(1) - u^{n+1}(0) = 0$, $u^{n+\tau_2}(1) - u^{n+\tau_2}(0) = 0$, and $u^{n+\tau_1}(1) - u^{n+\tau_1}(0) = 0$ correspond to the periodic boundary condition (\ref{eq:Advection_initial_boundary}). Moreover, $\bm{u}_3^n = \bm{u}_2^n = \bm{u}_1^n = \bm{u}^n$ and $\{\bm{x}^n,\bm{u}^n\}$ are the artificially generated data. This last equality reveals a key feature of this Runge-Kutta \emph{numerical Gaussian process}, namely the fact that it inspects the same data through the lens of different kernels.  A detailed derivation of the covariance matrix $\bm{K}$ is given in section \ref{Appendix:Advection} of the appendix. Prediction and propagation of uncertainty associated with the noisy initial observations can be performed as in section \ref{Appendix:Advection} of the appendix. Figure \ref{fig:Advection} depicts the noisy initial data along with the posterior distribution (\ref{eq:Advection_posterior}) of the solution to the advection equation (\ref{eq:Advection}) at different time snapshots.

\subsubsection{Numerical Study}
\begin{figure}
\centering
\includegraphics[width=\textwidth]{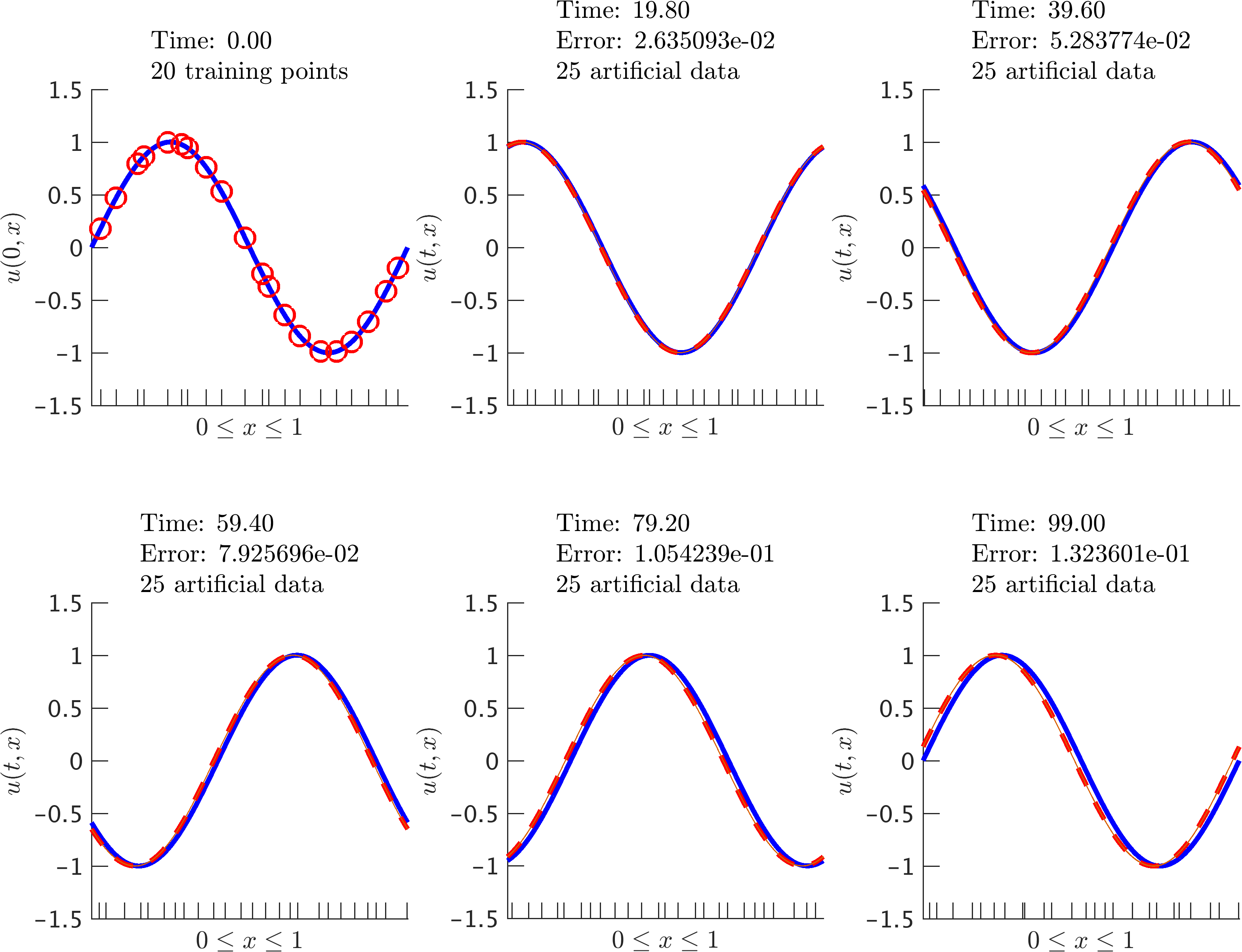}
\caption{\emph{Advection equation:} Initial data along with the posterior distribution of the solution at different time snapshots. The blue solid line represents the true data generating solution, while the dashed red line depicts the posterior mean. The shaded orange region illustrates the two standard deviations band around the mean. At each time step we generate $25$ artificial data points  randomly located in the interval $[0,1]$ according to a uniform distribution. These locations are highlighted by the ticks along the horizontal axis.  We are employing the Gauss-Legendre time-stepping quadrature with time step size $\Delta t = 0.1$. It is worth highlighting that we are running the time stepping scheme for a very long time with a relatively large time step size. We are reporting the relative spatial $\mathcal{L}^2$-error between the posterior mean and the true solution. {\it (Code: \url{http://bit.ly/2mpOtfQ}, Movie: \url{http://bit.ly/2m6XE2h})}}\label{fig:Advection_noiseless}
\end{figure}

\begin{figure}
\centering
\includegraphics[width=0.5\linewidth]{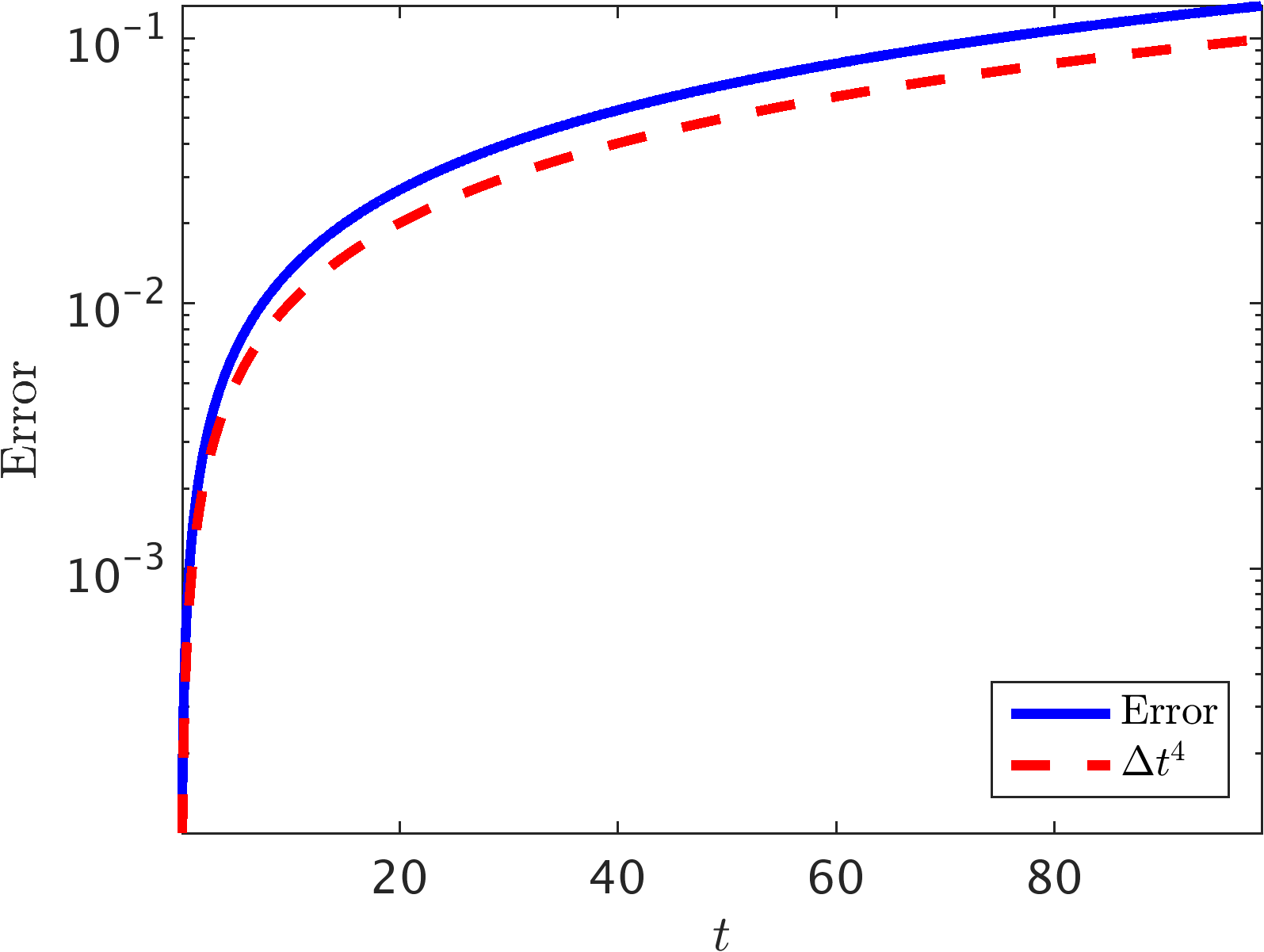}\\
\caption{\emph{Advection equation:} Time evolution of the relative spatial $\mathcal{L}^2$-error up to the final integration time $T=99.0$. We are using the Gauss-Legendre implicit Runge-Kutta scheme with a time step-size of $\Delta t = 0.1$. The red dashed line illustrates the optimal fourth-order convergence rate. {\it (Code: \url{http://bit.ly/2mntVDh})}} \label{fig:Advection_error_versus_time}
\vspace{5mm}
\minipage{0.48\textwidth}
  \includegraphics[width=\linewidth]{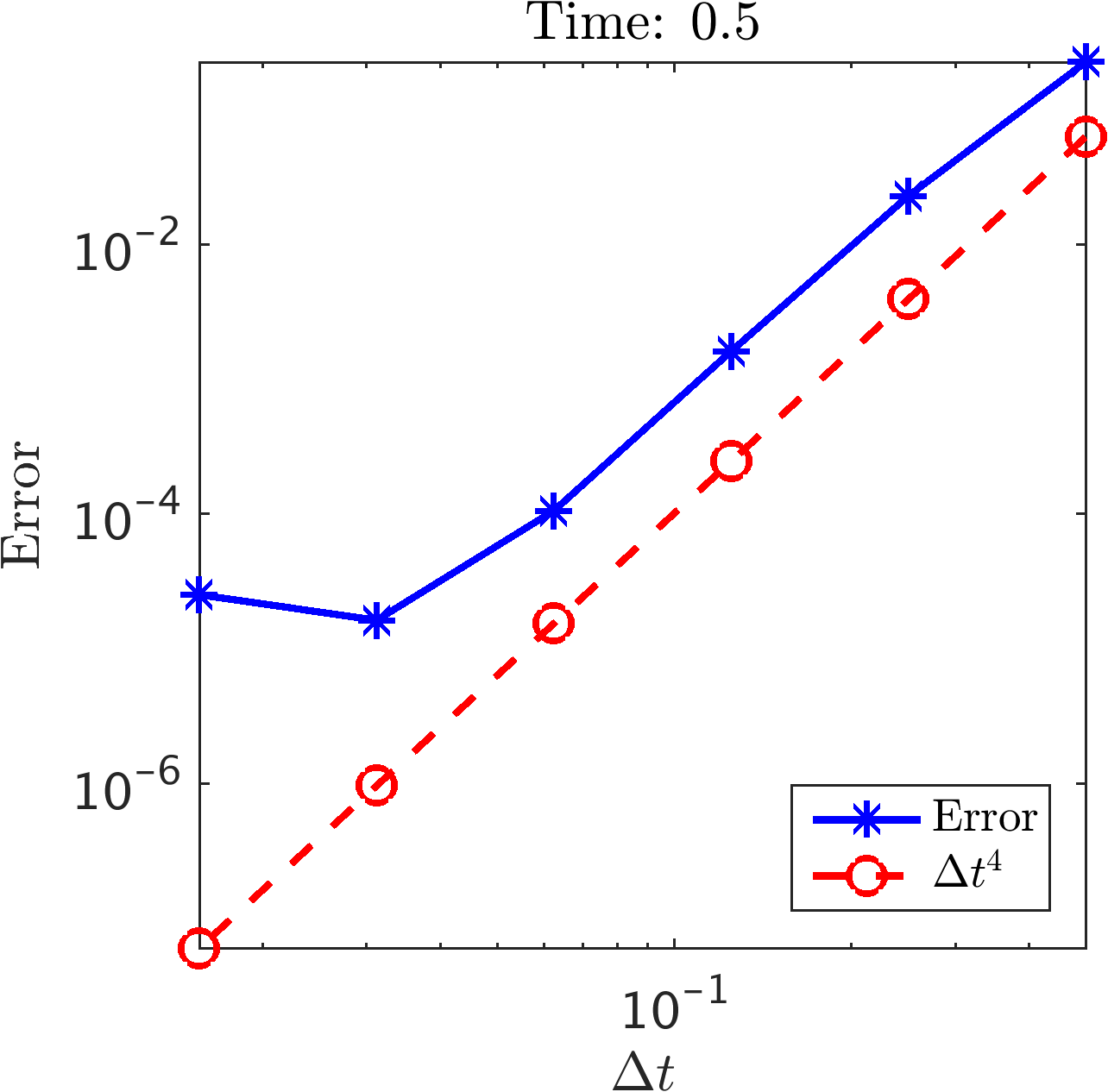}
  \caption{\emph{Advection equation:} Relative spatial $\mathcal{L}^2$-error versus step-size for the Gauss-Legendre method. Here, the number of noiseless initial data as well as the artificially generated data is set to be equal to $50$. We are running the time stepping scheme up until time $0.5$. {\it (Code: \url{http://bit.ly/2mntVDh})}}\label{fig:Advection_time_error}
\endminipage\hfill
\minipage{0.48\textwidth}
  \includegraphics[width=\linewidth]{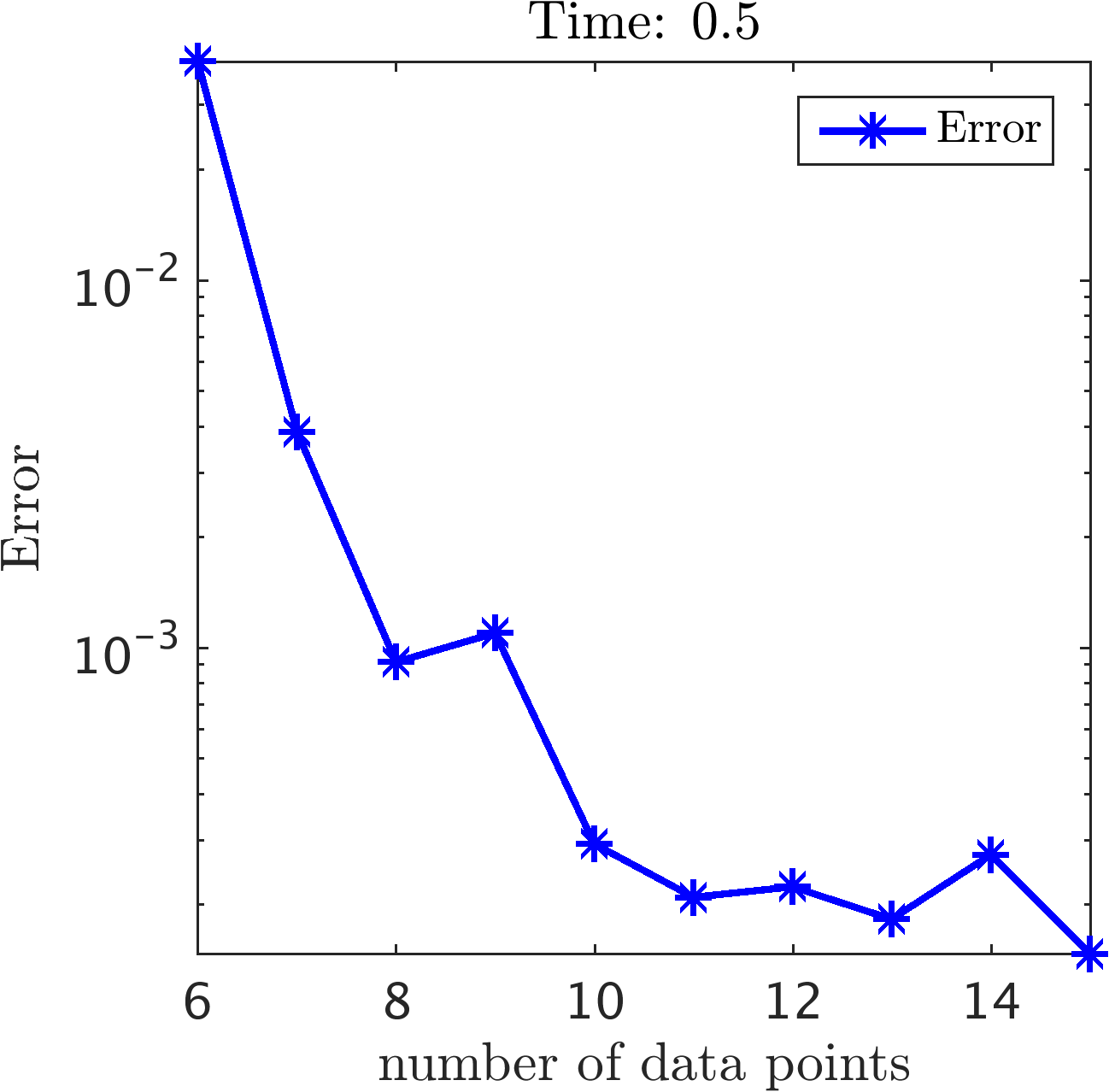}
  \caption{\emph{Advection equation:} Relative spatial $\mathcal{L}^2$-error versus the number of noiseless initial as well as artificial data points used for the Gauss-Legendre method. Here, the time step-size is set to be $\Delta t = 0.1$. We are running the time stepping scheme up until time $0.5$. {\it (Code: \url{http://bit.ly/2mntVDh})}}\label{fig:Advection_space_error}
\endminipage
\end{figure}

In the case where we have access to noiseless initial data we obtain the results depicted in figure \ref{fig:Advection_noiseless}.
Moreover, in order to make sure that the \emph{numerical Gaussian process} resulting from the Gauss-Legendre method (\ref{eq:GaussLegendre}) applied to the advection equation is indeed fourth-order accurate in time, we perform the numerical experiment reported in figures \ref{fig:Advection_error_versus_time} and \ref{fig:Advection_time_error}. The qualitative analysis of the temporal  as well as the spatial convergence properties (as seen in figure \ref{fig:Advection_space_error}) closely follows the conclusions drawn in section \ref{sec:Burgers_numerical_study}.

\subsection{Example: Heat Equation (Trapezoidal Rule)}\label{sec:Heat}
\begin{figure}
\centering
\includegraphics[width=\textwidth]{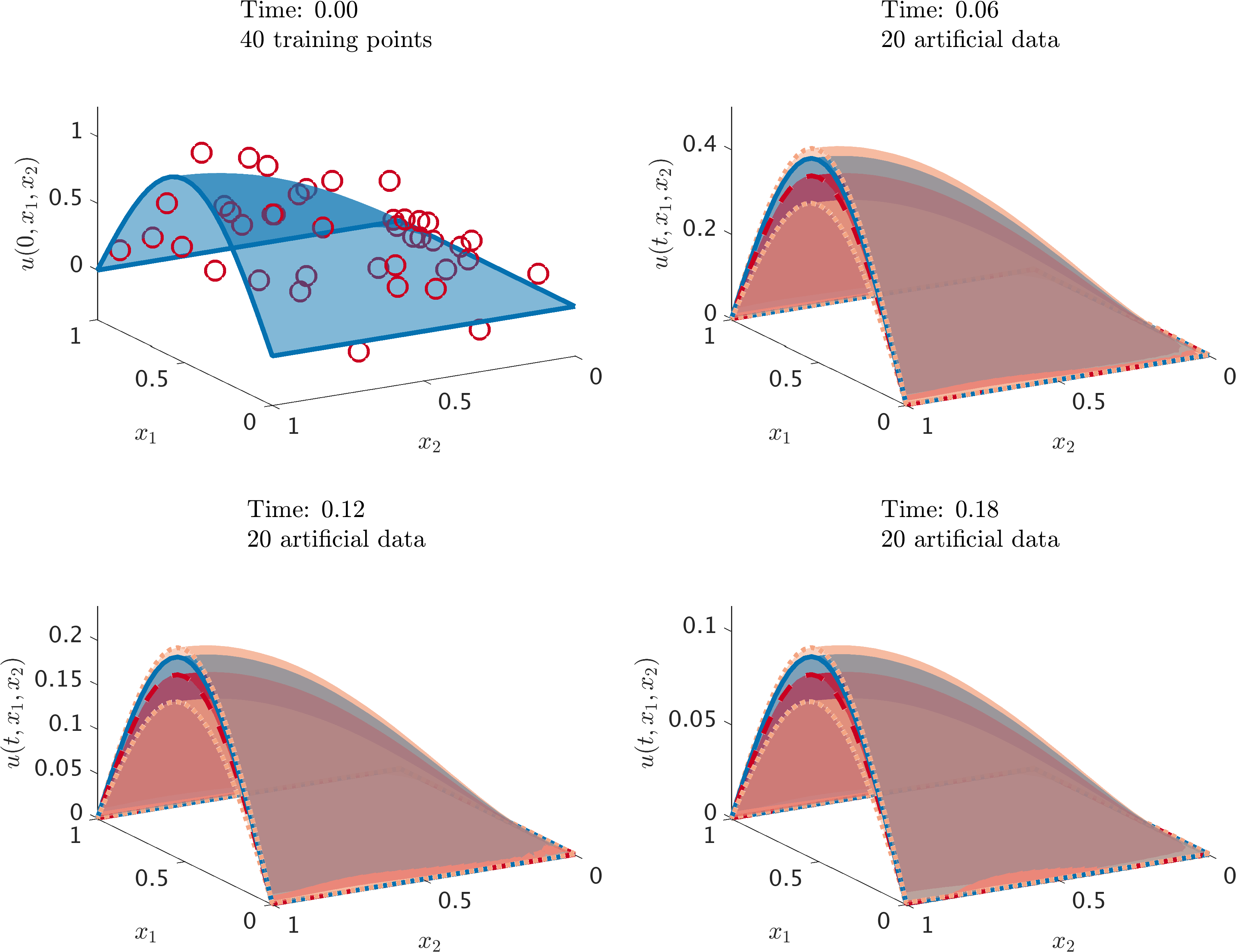}
\caption{\emph{Heat equation:} Initial data along with the posterior distribution of the solution at different time snapshots. The blue surface with solid lines represents the true data generating solution, while the red surface with dashed lines depicts the posterior mean. The two standard deviations band around the mean is depicted using the orange surface with dotted boundary. We are employing the trapezoidal rule with time step size $\Delta t = 0.01$. At each time step we generate $20$ artificial data points randomly located  in the domain $[0,1]\times[0,1]$ according to a uniform distribution. We employ three noiseless data-points per boundary. {\it (Code: \url{http://bit.ly/2mnFpGS}, Movie: \url{http://bit.ly/2mq4UZt})}}\label{fig:Heat}
\end{figure}

Revisiting the trapezoidal rule, equipped with the machinery introduced for the Runge-Kutta methods, we obtain an alternative \emph{numerical Gaussian process} to the one proposed in section \ref{sec:LinearMultistepMethods}. We will apply the resulting scheme to the heat equation in two space dimensions; i.e.,
\begin{equation}\label{eq:Heat}
u_t = u_{x_1x_1} + u_{x_2x_2},\ \ \ x_1 \in [0,1],\ x_2\in [0,1].
\end{equation}
The function $u(t,x_1,x_2) = e^{-\frac{5 \pi ^2 t}{4}} \sin (\pi  x_1) \sin \left(\frac{\pi  x_2}{2}\right)$ solves this equation and satisfies the following initial and boundary conditions
\begin{eqnarray}
&&u(0,x_1,x_2) = \sin (\pi  x_1) \sin \left(\frac{\pi  x_2}{2}\right),\nonumber\\
&&u(t,0,x_2) = u(t,1,x_2) = 0,\ \ u(t,x_1,0) = 0,\label{eq:Heat_Dirichlet_boundary}\\
&&u_{x_2}(t,x_1,1) = 0.\label{eq:Heat_Neumann_boundary}
\end{eqnarray}
Equations (\ref{eq:Heat_Dirichlet_boundary}) involve Dirichlet boundary conditions while equation (\ref{eq:Heat_Neumann_boundary}) corresponds to a Neumann-type boundary. Let us assume that all we observe are noisy measurements $\{(\bm{x}_1^0,\bm{x}_2^0), \bm{u}^0\}$ of the \emph{black-box} initial function $u(0,x_1,x_2)$. Given such measurements, we would like to infer the latent scalar field $u(t,x_1,x_2)$ (i.e., the solution to the heat equation (\ref{eq:Heat})), while quantifying the uncertainty associated with the noisy initial data (see figure \ref{fig:Heat}).
This example showcases the ability of the proposed methods to handle multi-dimensional spatial domains and mixed boundary conditions (see equations (\ref{eq:Heat_Dirichlet_boundary}) and (\ref{eq:Heat_Neumann_boundary})). Let us apply the trapezoidal scheme to the heat equation (\ref{eq:Heat}). The trapezoidal rule for the heat equation is given by
\begin{eqnarray}\label{eq:Heat_trapezoidal}
u^{n+1} = u^n &+& \frac{1}{2}\Delta t \frac{d^2}{d {x_1}^2}u^n + \frac{1}{2}\Delta t \frac{d^2}{d {x_1}^2}u^{n+1}\\
&+& \frac{1}{2}\Delta t \frac{d^2}{d {x_2}^2}u^n + \frac{1}{2}\Delta t \frac{d^2}{d {x_2}^2}u^{n+1}.\nonumber
\end{eqnarray}
Rearranging the terms, we can write $u^n_1 := u^n$ and
\begin{eqnarray}
u^{n}_2 := u^{n}_3 := u^{n} = u^{n+1} &-& \frac{1}{2}\Delta t \frac{d^2}{d x_1^2}u^n - \frac{1}{2}\Delta t \frac{d^2}{d x_1^2}u^{n+1}\\
&-& \frac{1}{2}\Delta t \frac{d^2}{d x_2^2}u^n - \frac{1}{2}\Delta t \frac{d^2}{d x_2^2}u^{n+1}.\nonumber
\end{eqnarray}
In other words, we are just rewriting equations (\ref{eq:GaussLegendre}) for the heat equation (\ref{eq:Heat}) with $\tau_1 = 0$, $\tau_2 = 1$, $b_1 = b_2 = \frac{1}{2}$, $a_{11} = a_{12} = 0$, and $a_{21} =  a_{22} = 1/2$.

\subsubsection{Prior}
Similar to the strategy (\ref{eq:RungeKutta_prior_assumption}) adopted for the Runge-Kutta methods, and as an alternative to the scheme used in section \ref{sec:LinearMultistepMethods}, we make the following prior assumptions:
\begin{eqnarray}
u^{n+1}(x_1,x_2) &\sim & \mathcal{GP}(0,k^{n+1,n+1}_{u,u}((x_1,x_2),(x_1',x_2');\theta_{n+1})),\\
u^{n}(x_1,x_2) &\sim & \mathcal{GP}(0,k^{n,n}_{u,u}((x_1,x_2),(x_1',x_2');\theta_{n})).\nonumber
\end{eqnarray}
Here, we employ anisotropic squared exponential covariance functions of the form
\begin{eqnarray*}
&&k^{n+1,n+1}_{u,u}((x_1,x_2),(x_1',x_2');\theta_{n+1})\\ 
&&=  \gamma_{n+1}^2 \exp\left(-\frac12 w_{n+1,1}(x_1 - x_1')^2 -\frac12 w_{n+1,2}(x_2 - x_2')^2\right),\\
&&k^{n,n}_{u,u}((x_1,x_2),(x_1',x_2');\theta_n)\\ 
&&=  \gamma_{n}^2 \exp\left(-\frac12 w_{n,1}(x_1 - x_1')^2 -\frac12 w_{n,2}(x_2 - x_2')^2\right).
\end{eqnarray*}
The hyper-parameters are given by $\theta_{n+1} = (\gamma_{n+1}^2, w_{n+1,1}, w_{n+1,2})$ and $\theta_{n} = (\gamma_{n}^2, w_{n,1}, w_{n,2})$. To deal with the mixed boundary conditions (\ref{eq:Heat_Dirichlet_boundary}) and (\ref{eq:Heat_Neumann_boundary}), let us define $v^{n+1} := \frac{d}{d x_2}u^{n+1}$ and $v^{n} := \frac{d}{d x_2}u^{n}$. We obtain the following \emph{numerical Gaussian process}
\begin{eqnarray*}
\left[\begin{array}{c}
u^{n+1}\\
v^{n+1}\\
u^n\\
v^n\\
u^n_3\\
u^n_1
\end{array}\right] \sim \mathcal{GP}\left(0,\left[\begin{array}{cccccc}
k^{n+1,n+1}_{u,u} & k^{n+1,n+1}_{u,v} & 0 & 0 & k^{n+1,n}_{u,3} & 0\\
 & k^{n+1,n+1}_{v,v} & 0 & 0 & k^{n+1,n}_{v,3} & 0\\
 & & k^{n,n}_{u,u} & k^{n,n}_{u,v} & k^{n,n}_{u,3} & k^{n,n}_{u,1}\\
 & & & k^{n,n}_{v,v} & k^{n,n}_{v,3} & k^{n,n}_{v,1}\\
 & & & & k^{n,n}_{3,3} & k^{n,n}_{3,1}\\
 & & & &  & k^{n,n}_{1,1}\\
\end{array}\right]\right),
\end{eqnarray*}
where the covariance functions are given in section \ref{Appendix:Heat} of the appendix.

\subsubsection{Training}
The hyper-parameters $\theta_{n+1}$ and $\theta_n$ can be trained by minimizing the Negative Log Marginal Likelihood resulting from
\begin{equation}\label{eq:Heat_NLML}
\left[\begin{array}{c}
\bm{u}^{n+1}_{D}\\
\bm{v}^{n+1}_{N}\\
\bm{u}^{n}_{D}\\
\bm{v}^{n}_{N}\\
\bm{u}^n_3\\
\bm{u}^n_1
\end{array}\right] \sim \mathcal{N}\left(0, \bm{K}\right),
\end{equation}
where $\{(\bm{x}_{1,D}^{n+1},\bm{x}_{2,D}^{n+1}), \bm{u}^{n+1}_D\}$ and $\{(\bm{x}_{1,D}^{n},\bm{x}_{2,D}^{n}),\bm{u}^n_D\}$ denote the data on the Dirichlet (\ref{eq:Heat_Dirichlet_boundary}) portion of the boundary, while $$\{(\bm{x}_{1,N}^{n+1},\bm{x}_{2,N}^{n+1}),\bm{u}^{n+1}_N\}\ \text{and}\ \{(\bm{x}_{1,N}^{n},\bm{x}_{2,N}^{n}),\bm{u}^n_N\}$$ correspond to the Neumann (\ref{eq:Heat_Neumann_boundary}) boundary data. Moreover, $\bm{u}^n_1 = \bm{u}^n_3 = \bm{u}^n$ and $\{(\bm{x}^n_1,\bm{x}^n_2),\bm{u}^n\}$ are the artificially generated data. The exact form of the covariance matrix $\bm{K}$ is given in section \ref{Appendix:Heat} of the appendix. Prediction and propagation of uncertainty associated with the noisy initial observations can be performed as in section \ref{Appendix:Heat} of the appendix. Figure \ref{fig:Heat} depicts the noisy initial data along with the posterior distribution (\ref{eq:Heat_posterior}) of the solution to the Heat equation (\ref{eq:Heat}) at different time snapshots. 

\subsubsection{Numerical Study}
\begin{figure}
\centering
\includegraphics[width=\textwidth]{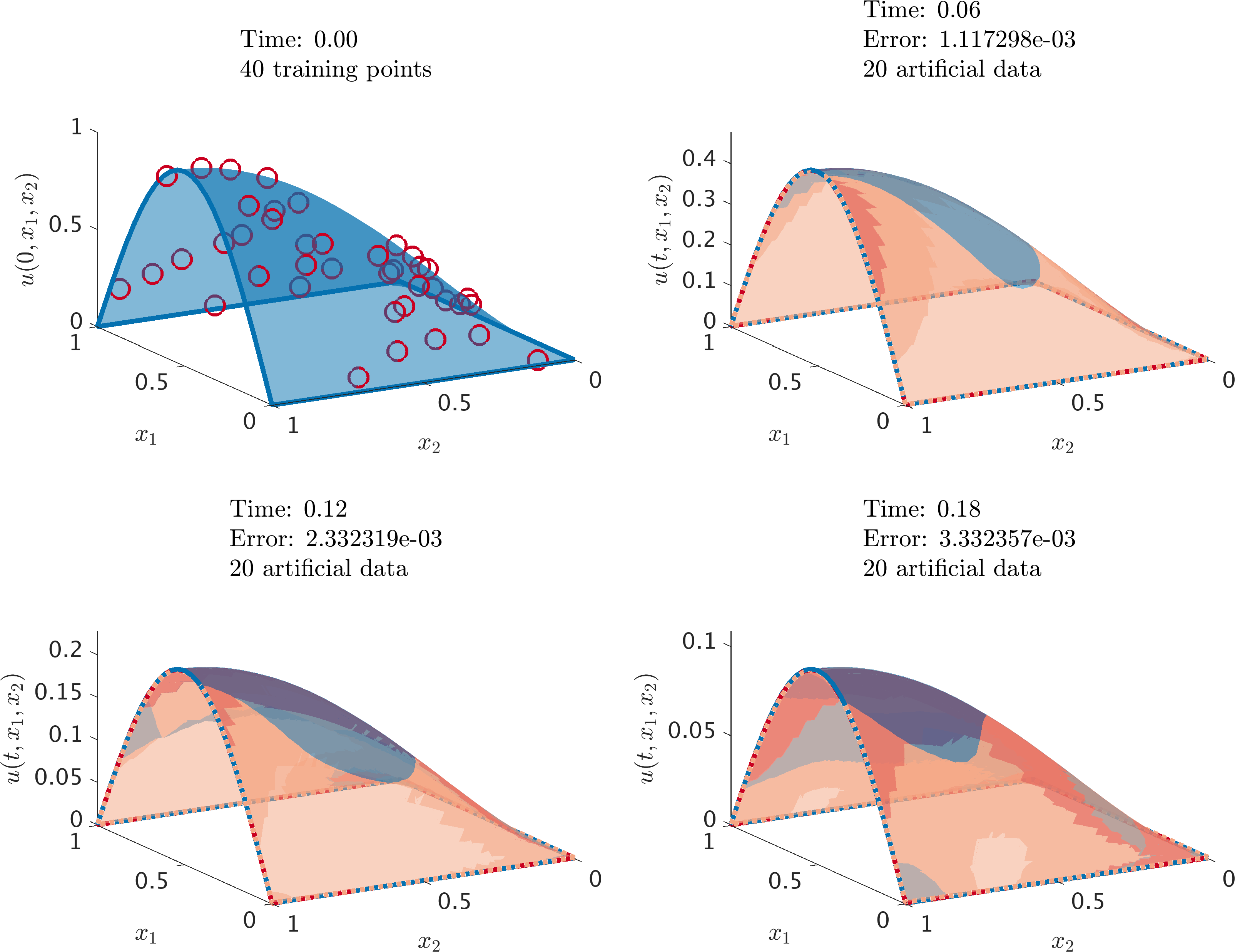}
\caption{\emph{Heat equation:} Initial data along with the posterior distribution of the solution at different time snapshots. The blue surface with solid lines represents the true data generating solution, while the red surface with dashed lines depicts the posterior mean. The two standard deviations band around the mean is depicted using the orange surface with dotted boundary. We are employing the trapezoidal rule with time step size $\Delta t = 0.01$. At each time step we generate $20$ artificial data points randomly located in the domain $[0,1]\times[0,1]$ according to a uniform distribution. We employ three noiseless data-points per boundary. We are reporting the relative $\mathcal{L}^2$-error between the posterior mean and the true solution. {\it (Code: \url{http://bit.ly/2mLwyB6}, Movie: \url{http://bit.ly/2mnFRod})}}\label{fig:Heat_noiseless}
\end{figure}

\begin{figure}
\centering
\includegraphics[width=0.5\linewidth]{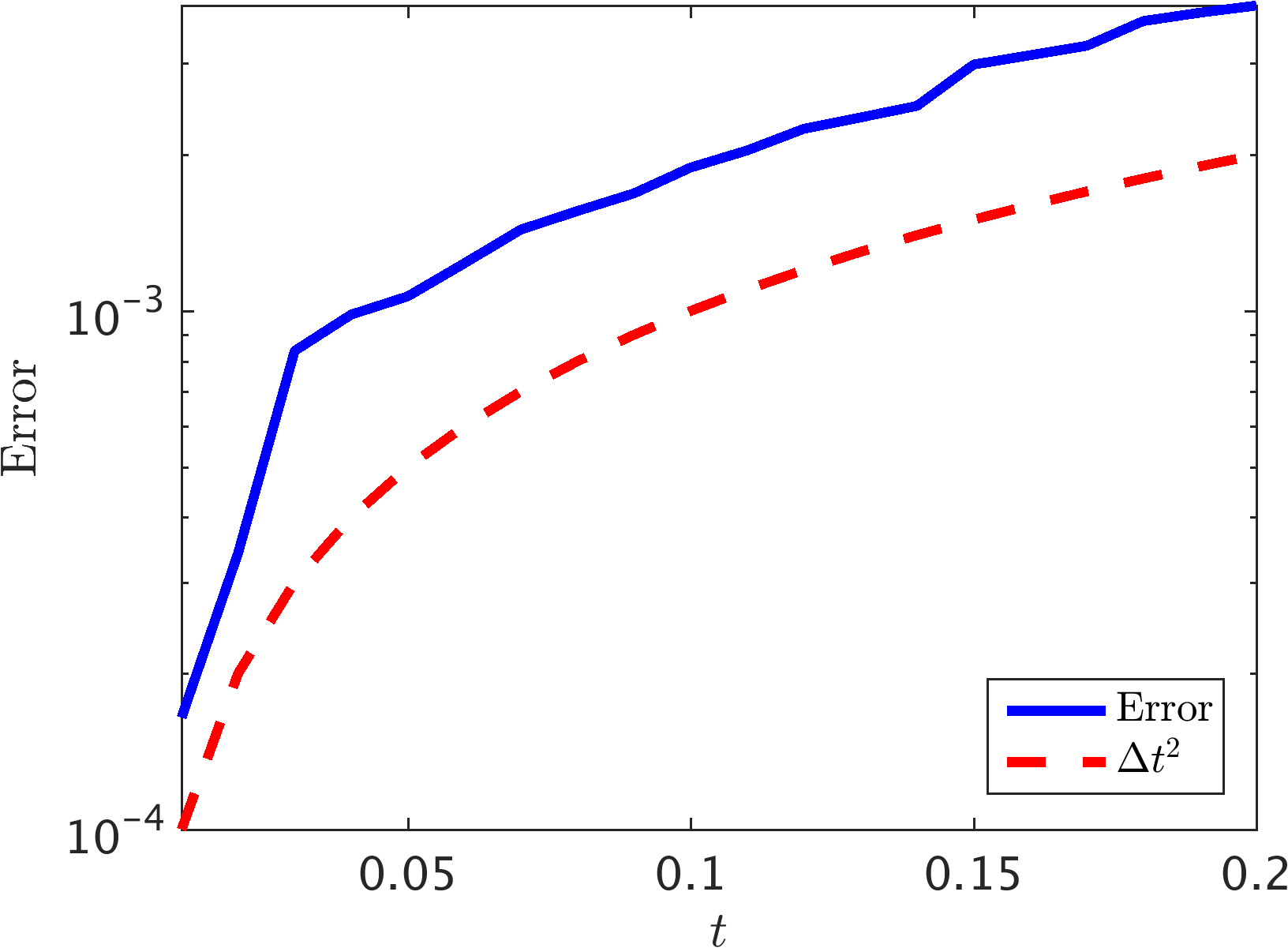}\\
\caption{\emph{Heat equation:} Time evolution of the relative spatial $\mathcal{L}^2$-error up to the final integration time $T=0.2$. We are using the trapezoidal rule with a time step-size of $\Delta t = 0.01$, and the red dashed line illustrates the optimal second-order convergence rate. {\it (Code: \url{http://bit.ly/2m7aoG9})}}\label{fig:Heat_error_versus_time}
\vspace{5mm}
\minipage{0.47\textwidth}
  \includegraphics[width=\linewidth]{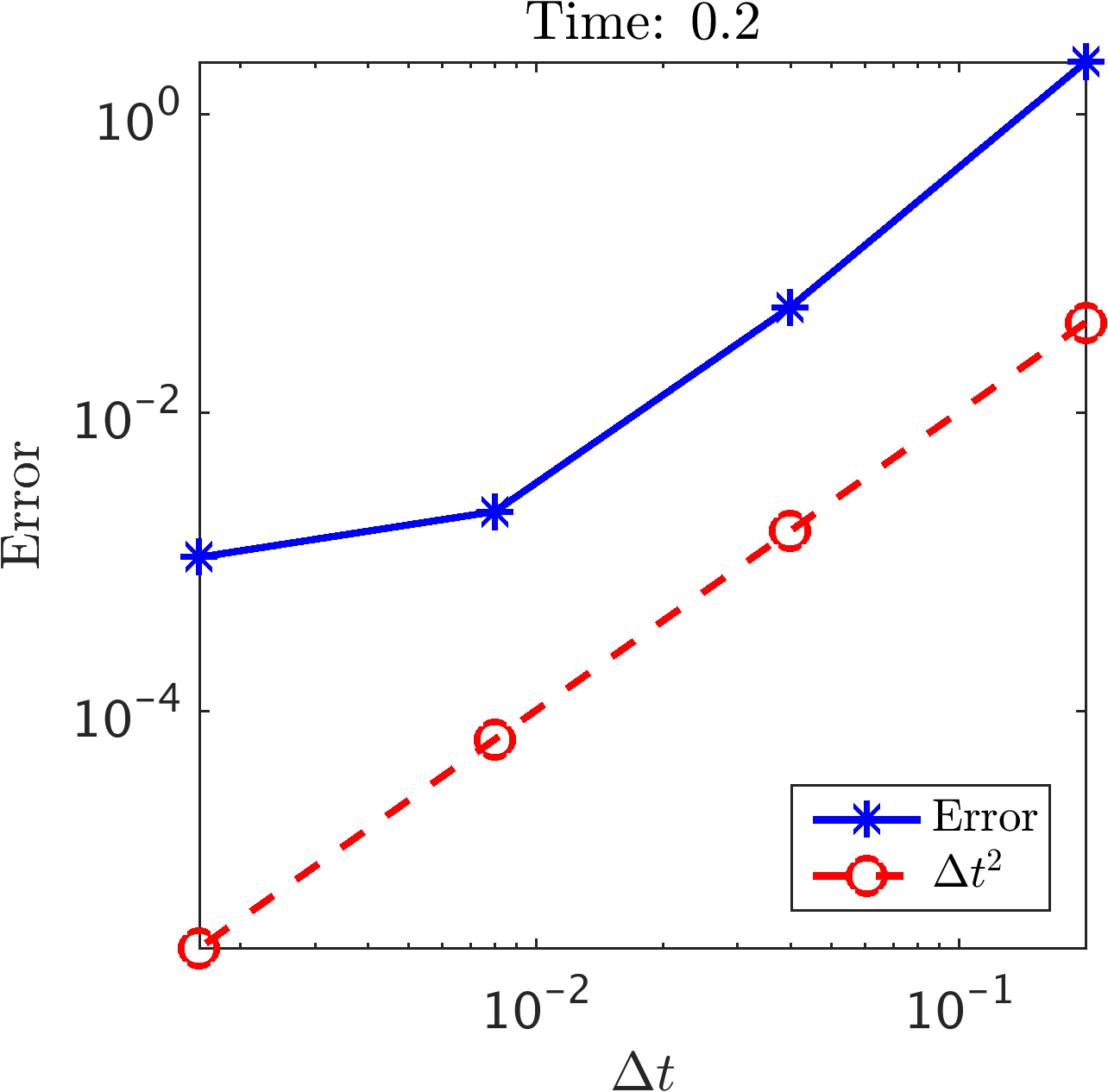}
  \caption{\emph{Heat equation:} Relative spatial $\mathcal{L}^2$-error versus step-size for the Runge-Kutta version of the trapezoidal rule at time $T = 0.2$. Here, the number of noiseless initial data as well as the artificially generated data is set to be equal to $50$. We are running the time stepping scheme up until time $0.2$. We employ $10$ noiseless data per boundary. {\it (Code: \url{http://bit.ly/2m7aoG9})}}\label{fig:Heat_time_error}
\endminipage\hfill
\minipage{0.47\textwidth}
  \includegraphics[width=\linewidth]{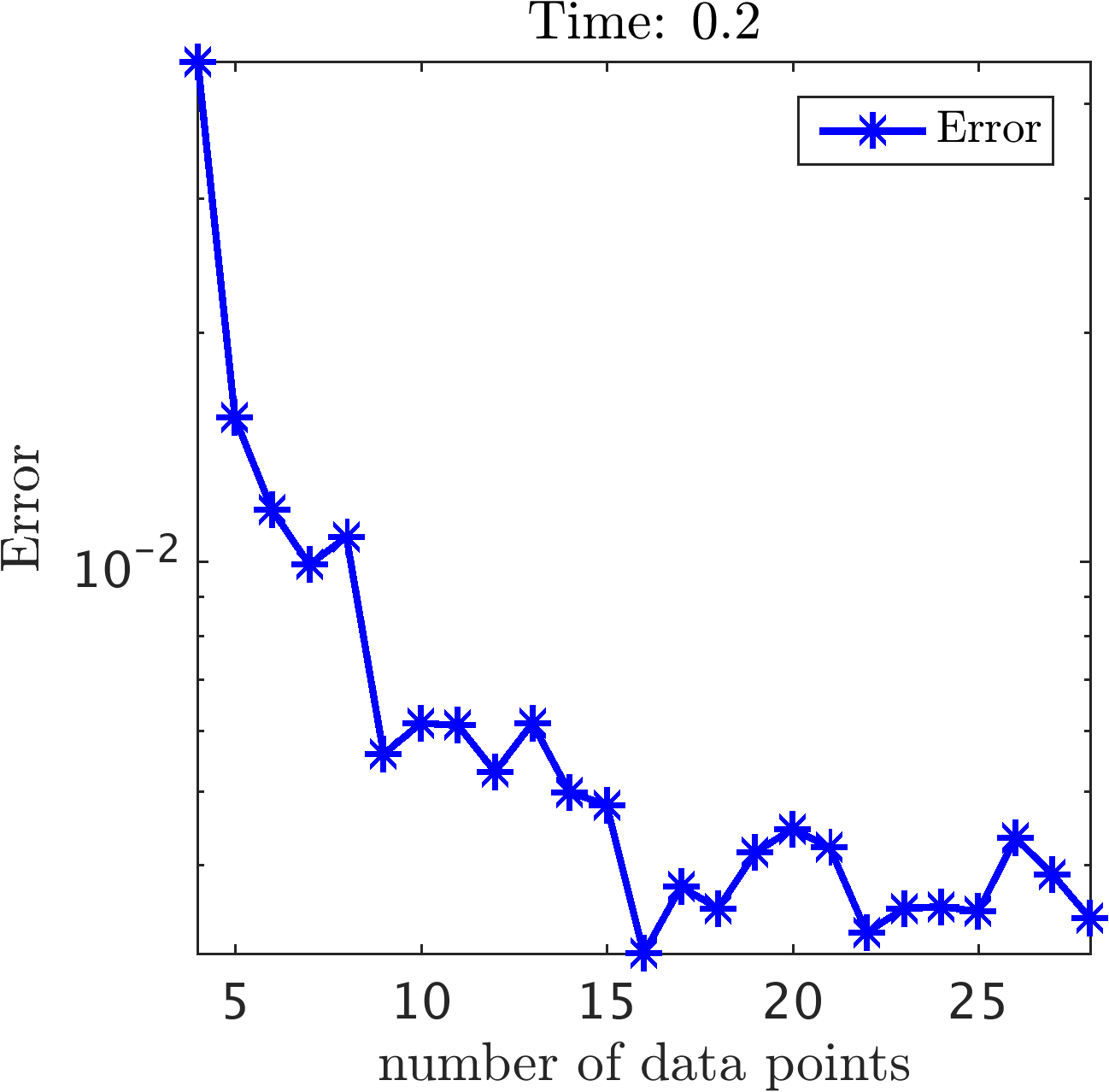}
  \caption{\emph{Heat equation:} Relative spatial $\mathcal{L}^2$-error versus the number of noiseless initial as well as artificial data points used for the Runge-Kutta version of the trapezoidal rule at time $T=0.2$. Here, the time step-size is set to be $\Delta t = 0.01$. We are running the time stepping scheme up until time $0.2$. We employ $10$ noiseless data per boundary. {\it (Code: \url{http://bit.ly/2m7aoG9})}}\label{fig:Heat_space_error}
\endminipage
\end{figure}

In order to be able to perform a systematic numerical study of the proposed methodology, we will operate under the assumption that we have access to \emph{noiseless} initial data. The corresponding results are reported in figure \ref{fig:Heat_noiseless}.
Moreover, in order to make sure that the \emph{numerical Gaussian process} resulting from the Runge-Kutta version of the trapezoidal rule (\ref{eq:Heat_trapezoidal}) applied to the Heat equation is indeed second-order accurate in time, we perform the numerical experiments reported in figures \ref{fig:Heat_error_versus_time} and \ref{fig:Heat_time_error}. Again, the qualitative analysis of the temporal  as well as the spatial convergence properties (as seen in figure \ref{fig:Heat_space_error}) closely follows the conclusions drawn in section \ref{sec:Burgers_numerical_study}.

\section{Concluding Remarks}

We have presented a novel machine learning framework for encoding physical laws described by partial differential equations into Gaussian process priors for nonparametric Bayesian regression. The proposed algorithms can be used to infer solutions to time-dependent and nonlinear partial differential equations, and effectively quantify and propagate uncertainty due to noisy initial or boundary data. Moreover, to the best of our knowledge, this is the first attempt to construct structured learning machines which are explicitly informed by the underlying physics that possibly generated the observed data. Exploiting this structure is critical for constructing data-efficient learning algorithms that can effectively distill information in the data-scarce scenarios appearing routinely when we study complex physical systems.

In contrast to classical deterministic numerical methods for solving partial differential equations (e.g., finite difference and finite-element methods), the proposed approach is by construction capable of propagating entire probability distributions in time. Although this provides a natural platform for learning from noisy data and computing under uncertainty, it comes with a non-negligible computational cost. Specifically, a limitation of this work in its present form stems from the cubic scaling with respect to the total number of training data points. In future work we plan to design more computationally efficient algorithms by exploring ideas including recursive Kalman updates \cite{hartikainen2010kalman} and variational inference \cite{hensman2013gaussian}.
 
From a classical numerical analysis standpoint, it also becomes natural to ask questions on convergence, derivation of dispersion relations, quantification of truncation errors, comparison against classical schemes, etc. We must underline that these questions  become obsolete in presence of noisy data and cannot be straightforwardly tackled using standard techniques from numerical analysis due to the probabilistic nature of the proposed work flow. In the realm of \emph{numerical Gaussian processes} such questions translate into investigating theoretical concepts like prior consistency \cite{Rasmussen06gaussianprocesses}, posterior robustness \cite{owhadi2015brittleness}, and posterior contraction rates \cite{stuart2016posterior}. These define a vast territory for analysis and future developments that currently remains unexplored.

In terms of future work, we plan to leverage the proposed framework to study more complex physical systems (e.g., fluid flows via the Navier-Stokes prior), propose extensions that can accommodate parameter inference, inverse and model discovery problems \cite{raissi2017machine}, as well as incorporate probabilistic time integration schemes that allow for a natural quantification of uncertainty due to time-stepping errors \cite{schober2014probabilistic}.

\section*{Acknowledgements}
This works received support by the DARPA EQUiPS grant N66001-15-2-4055, and the AFOSR grant 5210009.



%% References


\section*{References}
\bibliographystyle{model1-num-names}
\bibliography{sample.bib}






% \end{thebibliography}

\pagebreak
\section{Appendix}
\subsection{Burgers' Equation}\label{Appendix:Burgers}
The covariance functions for the Burgers' equation example are given by  $k^{n,n}_{u,u} = k$,
\begin{eqnarray}
k^{n,n-1}_{u,u} &=& k + \Delta t \mu^{n-1}(x')\frac{d}{d x'}k - \nu \Delta t \frac{d^2}{d x'^2} k,
\end{eqnarray}
and
\begin{eqnarray}
k^{n-1,n-1}_{u,u} &=& k + \Delta t \mu^{n-1}(x') \frac{d}{d x'} k  - \nu \Delta t \frac{d^2}{d x'^2} k,\\
&+& \Delta t \mu^{n-1}(x) \frac{d}{d x} k + \Delta t^2 \mu^{n-1}(x) \mu^{n-1}(x') \frac{d}{d x} \frac{d}{d x'} k\nonumber\\
&-& \nu \Delta t^2 \mu^{n-1}(x) \frac{d}{d x} \frac{d^2}{d x'^2} k - \nu \Delta t \frac{d^2}{d x^2} k\nonumber\\
&-& \nu \Delta t^2 \mu^{n-1}(x')\frac{d^2}{d x^2}\frac{d}{d x'} k + \nu^2 \Delta t^2 \frac{d^2}{d x^2}\frac{d^2}{d x'^2}k.\nonumber
\end{eqnarray}
The only non-trivial operations in the aforementioned kernel computations are the ones involving derivatives of the kernels which can be performed using any mathematical symbolic computation program like Wolfram Mathematica.

\subsection{Wave Equation}\label{Appendix:Wave}
\subsubsection{Prior}
The covariance functions for the wave equation example are given by
\begin{equation}\label{eq:Wave_trapezoidal_kernels}
\begin{array}{ll}
k^{n,n}_{u,u} = k_u + \frac{1}{4} \Delta t^2 k_v, & k^{n,n}_{u,v} = \frac12 \Delta t \frac{d^2}{d x'^2} k_{u} + \frac12 \Delta t k_v,\\
k^{n,n-1}_{u,u} = k_u - \frac{1}{4} \Delta t^2 k_v, & k^{n,n-1}_{u,v} = -\frac12 \Delta t \frac{d^2}{d x'^2} k_{u} + \frac12 \Delta t k_v,\\
k^{n,n}_{v,v} =  k_{v} + \frac14 \Delta t^2 \frac{d^2}{d x^2}\frac{d^2}{d x'^2}k_u, & k^{n,n-1}_{v,u} =  -\frac12 \Delta t k_v + \frac12 \Delta t \frac{d^2}{d x^2} k_u,\\
k^{n,n-1}_{v,v} =  k_{v} - \frac14 \Delta t^2 \frac{d^2}{d x^2}\frac{d^2}{d x'^2}k_u, & k^{n-1,n-1}_{u,u} = k_u + \frac{1}{4} \Delta t^2 k_v,\\
k^{n-1,n-1}_{u,v} = -\frac12 \Delta t \frac{d^2}{d x'^2} k_{u} - \frac12 \Delta t k_v, & k^{n-1,n-1}_{v,v} =  k_{v} + \frac14 \Delta t^2 \frac{d^2}{d x^2}\frac{d^2}{d x'^2}k_u.
\end{array}
\end{equation}
It is worth highlighting that the only non-trivial but straightforward operations involved in the aforementioned kernel computations are
\begin{eqnarray}
\frac{d^2}{d x'^2} k_{u}(x,x';\theta_u) &=& \frac{d^2}{d x^2} k_{u}(x,x';\theta_u)\\
 &=& \gamma_u^2 w_u e^{-\frac{1}{2} w_u (x-x')^2} \left(w_u (x-x')^2-1\right),\nonumber\\
\frac{d^2}{d x^2}\frac{d^2}{d x'^2}k_u(x,x';\theta_u) &=& \gamma_u^2 w_u^2 e^{-\frac{1}{2} w_u (x-x')^2} \left(w_u (x-x')^2 \left(w_u
   (x-x')^2-6\right)+3\right) .\nonumber
\end{eqnarray}

\subsubsection{Training}
The hyper-parameters $\theta_u$ and $\theta_v$ can be trained by minimizing the Negative Log Marginal Likelihood resulting from
\begin{equation}\label{eq:Wave_NLML}
\left[\begin{array}{c}
\bm{u}^{n}_b \\ 
\bm{u}^{n-1} \\
\bm{v}^{n-1}
\end{array} \right] \sim \mathcal{N}\left(0,\bm{K}\right),
\end{equation}
where $\{\bm{x}^{n}_b, \bm{u}^{n}_b\}$ are the data on the boundary, $\{\bm{x}^{n-1}_u, \bm{u}^{n-1}\}$, $\{\bm{x}^{n-1}_v, \bm{v}^{n-1}\}$  are \emph{artificially generated data}, and
\[
\bm{K} := \left[\begin{array}{ccc}
\bm{K}^{n,n}_{u,u} + \sigma_{n}^2 I & \bm{K}^{n,n-1}_{u,u} & \bm{K}^{n,n-1}_{u,v}\\ 
 & \bm{K}^{n-1,n-1}_{u,u} + \sigma^2_{u, n-1}I & \bm{K}^{n-1,n-1}_{u,v} \\
 &   & \bm{K}^{n-1,n-1}_{v,v} + \sigma^2_{v,n-1}I
\end{array} \right],
\]
where
\begin{equation*}
\begin{array}{ll}
\bm{K}^{n,n}_{u,u} = k^{n,n}_{u,u}(\bm{x}_b^{n},\bm{x}_b^{n}), & \bm{K}^{n,n-1}_{u,u} = k^{n,n-1}_{u,u}(\bm{x}_b^{n},\bm{x}^{n-1}_u),\\
\bm{K}^{n,n-1}_{u,v} = k^{n,n-1}_{u,v}(\bm{x}_b^{n},\bm{x}_v^{n-1}), & \bm{K}^{n-1,n-1}_{u,u} = k^{n-1,n-1}_{u,u}(\bm{x}^{n-1}_u,\bm{x}^{n-1}_u),\\
\bm{K}^{n-1,n-1}_{u,v} = k^{n-1,n-1}_{u,v}(\bm{x}^{n-1}_u,\bm{x}^{n-1}_v), & \bm{K}^{n-1,n-1}_{v,v} = k^{n-1,n-1}_{v,v}(\bm{x}^{n-1}_v,\bm{x}^{n-1}_v).
\end{array}
\end{equation*}
Here, the data on the boundary are given by
\[
\bm{x}_b^n = \left[\begin{array}{c}
0\\
1
\end{array}\right], \ \ \ \bm{u}_b^n = \left[\begin{array}{c}
0\\
0
\end{array}\right],
\]
which correspond to the Dirichlet boundary conditions (\ref{eq:Wave_initial_boundary}).

\subsubsection{Posterior}
In order to predict $u^{n}(x^{n}_{*u})$ and $v^{n}(x^{n}_{*v})$ at new test points $x^{n}_{*u}$ and $x^{n}_{*v}$, respectively, we use the following conditional distribution
\begin{eqnarray*}
&&\left[\begin{array}{c}
u^{n}(x^{n}_{*u})\\
v^{n}(x^{n}_{*v})\\
\end{array}\right]
\ | \left[\begin{array}{c}
\bm{u}^{n}_b \\ 
\bm{u}^{n-1} \\
\bm{v}^{n-1}
\end{array} \right] \sim \\
&&\mathcal{N}\left(
\bm{q}^T
 \bm{K}^{-1}\left[\begin{array}{c}
\bm{u}^{n}_b \\ 
\bm{u}^{n-1}\\
\bm{v}^{n-1}
\end{array} \right], \left[\begin{array}{ccc}
k^{n,n}_{u,u}(x^{n}_{*u},x^{n}_{*u})  & k^{n,n}_{u,v}(x^{n}_{*u},x^{n}_{*v})  \\
  & k^{n,n}_{v,v}(x^{n}_{*v},x^{n}_{*v})  \\
\end{array} \right]  -  \bm{q}^T\bm{K}^{-1}\bm{q}\right),
\end{eqnarray*}
where $\bm{q} = [\bm{q}_u\ \bm{q}_v]$ and 
\begin{eqnarray*}
\bm{q}^T_u &:=& \left[\begin{array}{ccc}
k^{n,n}_{u,u}(x_{*u}^{n},\bm{x}^{n}_b)  & k^{n,n-1}_{u,u}(x_{*u}^{n},\bm{x}_u^{n-1}) & k^{n,n-1}_{u,v}(x_{*u}^{n},\bm{x}_v^{n-1})
\end{array} \right],\\
\bm{q}^T_v &:=& \left[\begin{array}{ccc}
k^{n,n}_{v,u}(x_{*v}^{n},\bm{x}^{n}_b)  & k^{n,n-1}_{v,u}(x_{*v}^{n},\bm{x}_u^{n-1}) & k^{n,n-1}_{v,v}(x_{*v}^{n},\bm{x}_v^{n-1})
\end{array} \right].
\end{eqnarray*}

\subsubsection{Propagating Uncertainty}
Since $\{\bm{x}^{n-1}_u, \bm{u}^{n-1}\}$ and $\{\bm{x}^{n-1}_v, \bm{v}^{n-1}\}$ are \emph{artificially generated data}, to properly propagate the uncertainty associated with the initial data, we have to marginalize them out by employing
\[
\left[\begin{array}{c}
\bm{u}^{n-1}\\
\bm{v}^{n-1}
\end{array}\right] \sim \mathcal{N}\left(
\left[\begin{array}{c}
\bm{\mu}_u^{n-1}\\
\bm{\mu}_v^{n-1}
\end{array}\right],\left[\begin{array}{cc}
\bm{\Sigma}^{n-1,n-1}_{u,u} & \bm{\Sigma}^{n-1,n-1}_{u,v}\\
 & \bm{\Sigma}^{n-1,n-1}_{v,v}
\end{array}\right]\right),
\]
to obtain
\begin{eqnarray}\label{eq:Wave_posterior}
&&\left[\begin{array}{c}
u^{n}(x^{n}_{*u})\\
v^{n}(x^{n}_{*v})\\
\end{array}\right]\ |\ \bm{u}^{n}_b
\sim \\
&&\mathcal{N}\left(\left[\begin{array}{c}
\mu^{n}_u(x_{*u}^{n})\\
\mu^{n}_v(x_{*v}^{n})
\end{array}\right],  \left[\begin{array}{cc}
\Sigma^{n,n}_{u,u}(x_{*u}^{n},x_{*u}^{n}) & \Sigma^{n,n}_{u,v}(x_{*u}^{n},x_{*v}^{n})\\
 & \Sigma^{n,n}_{v,v}(x_{*v}^{n},x_{*v}^{n}) 
\end{array}\right]
\right),\nonumber
\end{eqnarray}
where
\[
\left[\begin{array}{c}
\mu^{n}_u(x_{*u}^{n})\\
\mu^{n}_v(x_{*v}^{n})
\end{array}\right] = \bm{q}^T \bm{K}^{-1}\left[\begin{array}{c}
\bm{u}^{n}_b \\ 
\bm{\mu}^{n-1}_u\\
\bm{\mu}^{n-1}_v
\end{array} \right],
\]
and
\begin{eqnarray*}
&&\left[\begin{array}{cc}
\Sigma^{n,n}_{u,u}(x_{*u}^{n},x_{*u}^{n}) & \Sigma^{n,n}_{u,v}(x_{*u}^{n},x_{*v}^{n})\\
 & \Sigma^{n,n}_{v,v}(x_{*v}^{n},x_{*v}^{n}) 
\end{array}\right] =\left[\begin{array}{ccc}
k^{n,n}_{u,u}(x^{n}_{*u},x^{n}_{*u})  & k^{n,n}_{u,v}(x^{n}_{*u},x^{n}_{*v})  \\
  & k^{n,n}_{v,v}(x^{n}_{*v},x^{n}_{*v})  \\
\end{array} \right] - \\
&&  \bm{q}^T\bm{K}^{-1}\bm{q} + \bm{q}^T\bm{K}^{-1} \left[\begin{array}{ccc}
0 & 0  & 0\\ 
 & \bm{\Sigma}^{n-1,n-1}_{u,u} & \bm{\Sigma}^{n-1,n-1}_{u,v}\\
&  & \bm{\Sigma}^{n-1,n-1}_{v,v} 
\end{array} \right]\bm{K}^{-1}\bm{q}.
\end{eqnarray*}
Now, we can use the resulting posterior distribution to obtain the artificially generated data $\{\bm{x}^{n}_u,\bm{u}^{n}\}$ and $\{\bm{x}^{n}_v,\bm{v}^{n}\}$ for the next time step with
\begin{equation}\label{eq:Wave_artificial_data}
\left[\begin{array}{c}
\bm{u}^{n}\\
\bm{v}^{n}
\end{array}\right] \sim \mathcal{N}\left(
\left[\begin{array}{c}
\bm{\mu}_u^{n}\\
\bm{\mu}_v^{n}
\end{array}\right],\left[\begin{array}{cc}
\bm{\Sigma}^{n,n}_{u,u} & \bm{\Sigma}^{n,n}_{u,v}\\
 & \bm{\Sigma}^{n,n}_{v,v}
\end{array}\right]\right).
\end{equation}

\subsection{Advection Equation}\label{Appendix:Advection}
\subsubsection{Prior}
The covariance functions for the advection equation example are given by
\begin{equation*}\label{eq:Advection_Kernels}
\begin{array}{ll}
k^{n+1,n}_{u,3} = k^{n+1,n+1}_{u,u}, & k^{n+\tau_2,n}_{u,3} = b_2 \Delta t  \frac{d}{d x'}k^{n+\tau_2,n+\tau_2}_{u,u},\\
k^{n+\tau_2,n}_{u,2} = k^{n+\tau_2,n+\tau_2}_{u,u} + a_{22} \Delta t  \frac{d}{d x'}k^{n+\tau_2,n+\tau_2}_{u,u}, & k^{n+\tau_2,n}_{u,1} = a_{12} \Delta t  \frac{d}{d x'}k^{n+\tau_2,n+\tau_2}_{u,u},\\
k^{n+\tau_1,n}_{u,3} = b_1 \Delta t  \frac{d}{d x'}k^{n+\tau_1,n+\tau_1}_{u,u}, & k^{n+\tau_1,n}_{u,2} = a_{21} \Delta t  \frac{d}{d x'}k^{n+\tau_1,n+\tau_1}_{u,u},\\
k^{n+\tau_1,n}_{u,1} = k^{n+\tau_1,n+\tau_1}_{u,u} + a_{11} \Delta t  \frac{d}{d x'}k^{n+\tau_1,n+\tau_1}_{u,u}, & \\
\end{array}
\end{equation*}
and
\begin{eqnarray}
k^{n,n}_{3,3} &=& k^{n+1,n+1}_{u,u} + b_1^2 \Delta t^2  \frac{d}{d x} \frac{d}{d x'} k^{n+\tau_1,n+\tau_1}_{u,u} + b_2^2 \Delta t^2  \frac{d}{d x} \frac{d}{d x'} k^{n+\tau_2,n+\tau_2}_{u,u},\nonumber\\
k^{n,n}_{3,2} &=& b_2 \Delta t \frac{d}{d x} k^{n+\tau_2,n+\tau_2}_{u,u} + a_{21} b_1 \Delta t^2  \frac{d}{d x} \frac{d}{d x'} k^{n+\tau_1,n+\tau_1}_{u,u}\nonumber\\ &+& a_{22} b_2\Delta t^2  \frac{d}{d x} \frac{d}{d x'} k^{n+\tau_2,n+\tau_2}_{u,u},\nonumber\\
k^{n,n}_{3,1} &=& b_1 \Delta t \frac{d}{d x} k^{n+\tau_1,n+\tau_1}_{u,u} + a_{11} b_1 \Delta t^2  \frac{d}{d x} \frac{d}{d x'} k^{n+\tau_1,n+\tau_1}_{u,u} \nonumber\\ &+& a_{12} b_2\Delta t^2  \frac{d}{d x} \frac{d}{d x'} k^{n+\tau_2,n+\tau_2}_{u,u},\nonumber
\end{eqnarray}
\begin{eqnarray}
k^{n,n}_{2,2} &=& k^{n+\tau_2,n+\tau_2}_{u,u} + a_{21}^2 \Delta t^2  \frac{d}{d x}\frac{d}{d x'} k^{n+\tau_1,n+\tau_1}_{u,u} + a_{22}^2 \Delta t^2 \frac{d}{d x}\frac{d}{d x'} k^{n+\tau_2,n+\tau_2}_{u,u},\nonumber\\
k^{n,n}_{2,1} &=& a_{12} \Delta t \frac{d}{d x'} k^{n+\tau_2,n+\tau_2}_{u,u} + a_{21} \Delta t \frac{d}{d x} k^{n+\tau_1,n+\tau_1}_{u,u}\nonumber\\
&+& a_{21} a_{11} \Delta t^2   \frac{d}{d x}\frac{d}{d x'} k^{n+\tau_1,n+\tau_1}_{u,u} + a_{22}a_{12} \Delta t^2 \frac{d}{d x}\frac{d}{d x'} k^{n+\tau_2,n+\tau_2}_{u,u},\nonumber
\end{eqnarray}
\begin{eqnarray}
k^{n,n}_{1,1} &=& k^{n+\tau_1,n+\tau_1}_{u,u} + a_{11}^2 \Delta t^2  \frac{d}{d x}\frac{d}{d x'} k^{n+\tau_1,n+\tau_1}_{u,u} + a_{12}^2 \Delta t^2  \frac{d}{d x}\frac{d}{d x'} k^{n+\tau_2,n+\tau_2}_{u,u}.\nonumber
\end{eqnarray}

\subsubsection{Training}
The matrix $\bm{K}$ used in the distribution (\ref{eq:Advection_NLML}) is given by
\[
\resizebox{\textwidth}{!}{$
\bm{K} = \left[\begin{array}{cccccc}
K^{n+1,n+1}_{b,b} & 0 & 0 & \bm{K}^{n+1,n}_{b,3} & 0 & 0\\
 & K^{n+\tau_2,n+\tau_2}_{b,b} & 0  & \bm{K}^{n+\tau_2,n}_{b,3} & \bm{K}^{n+\tau_2,n}_{b,2} & \bm{K}^{n+\tau_2,n}_{b,1}\\
 &  & K^{n+\tau_1,n+\tau_1}_{b,b}  & \bm{K}^{n+\tau_1,n}_{b,3} & \bm{K}^{n+\tau_1,n}_{b,2} & \bm{K}^{n+\tau_1,n}_{b,1}\\
 & & &\bm{K}^{n,n}_{3,3}  + \sigma_n^2 I & \bm{K}^{n,n}_{3,2} & \bm{K}^{n,n}_{3,1}\\
 & & &  & \bm{K}^{n,n}_{2,2} + \sigma_n^2 I & \bm{K}^{n,n}_{2,1}\\
 & & &  &  & \bm{K}^{n,n}_{1,1} + \sigma_n^2 I\\
\end{array}\right],
$}
\]
where
\begin{eqnarray*}
K^{n+1,n+1}_{b,b} &=& k^{n+1,n+1}_{u,u}(1,1) - k^{n+1,n+1}_{u,u}(1,0)\nonumber\\ &-& k^{n+1,n+1}_{u,u}(0,1) + k^{n+1,n+1}_{u,u}(0,0),\\
K^{n+\tau_2,n+\tau_2}_{b,b} &=& k^{n+\tau_2,n+\tau_2}_{u,u}(1,1) - k^{n+\tau_2,n+\tau_2}_{u,u}(1,0)\nonumber\\ &-& k^{n+\tau_2,n+\tau_2}_{u,u}(0,1) + k^{n+\tau_2,n+\tau_2}_{u,u}(0,0),\\
K^{n+\tau_1,n+\tau_1}_{b,b} &=& k^{n+\tau_1,n+\tau_1}_{u,u}(1,1) - k^{n+\tau_1,n+\tau_1}_{u,u}(1,0)\nonumber\\ &-& k^{n+\tau_1,n+\tau_1}_{u,u}(0,1) + k^{n+\tau_1,n+\tau_1}_{u,u}(0,0),
\end{eqnarray*}
\begin{eqnarray*}
\bm{K}^{n+1,n}_{b,i} &=& k^{n+1,n}_{u,i}(1,\bm{x}^n) - k^{n+1,n}_{u,i}(0,\bm{x}^n),\ i=3,2,1,\\
\bm{K}^{n+\tau_2,n}_{b,i} &=& k^{n+\tau_2,n}_{u,i}(1,\bm{x}^n) - k^{n+\tau_2,n}_{u,i}(0,\bm{x}^n),\ i=3,2,1,\\
\bm{K}^{n+\tau_1,n}_{b,i} &=& k^{n+\tau_1,n}_{u,i}(1,\bm{x}^n) - k^{n+\tau_1,n}_{u,i}(0,\bm{x}^n),\ i=3,2,1,\\
\bm{K}^{n,n}_{i,j} &=& k^{n,n}_{i,j}(\bm{x}^n,\bm{x}^n), \ \ \ i,j=3,2,1, \ j \leq i.
\end{eqnarray*}

\subsubsection{Posterior}
In order to predict $u^{n+1}(x^{n+1}_*)$ at a new test point $x^{n+1}_*$, we use
\begin{eqnarray*}
&&u^{n+1}(x^{n+1}_*)\ | \left[\begin{array}{c}
u^{n+1}(1) - u^{n+1}(0) =0 \\ 
u^{n+\tau_2}(1) - u^{n+\tau_2}(0) =0 \\ 
u^{n+\tau_1}(1) - u^{n+\tau_1}(0) =0 \\ 
\bm{u}^n \\
\bm{u}^n \\
\bm{u}^{n}
\end{array} \right] \sim \\
&&\mathcal{N}\left(
\bm{q}^T \bm{K}^{-1}\left[\begin{array}{c}
0 \\ 
0 \\ 
0 \\ 
\bm{u}^n \\
\bm{u}^n \\
\bm{u}^{n}
\end{array} \right],k^{n+1,n+1}_{u,u}(x^{n+1}_*,x^{n+1}_*) - \bm{q}^T\bm{K}^{-1}\bm{q}\right),
\end{eqnarray*}
where
\[
\bm{q} := \left[\begin{array}{c}
k^{n+1,n+1}_{u,u}(1,x^{n+1}_*) - k^{n+1,n+1}_{u,u}(0,x^{n+1}_*)\\
0\\
0\\
k^{n,n+1}_{3,u}(\bm{x}^n,x_*^{n+1})\\ 
k^{n,n+1}_{2,u}(\bm{x}^n,x_*^{n+1})\\ 
k^{n,n+1}_{1,u}(\bm{x}^n,x_*^{n+1})\\ 
\end{array} \right].
\]

\subsubsection{Propagating Uncertainty}
To propagate the uncertainty associate with the noisy initial data through time we have to marginalize out the artificially generated data $\{\bm{x}^n, \bm{u}^n\}$ by employing
\[
\bm{u}^n \sim \mathcal{N}\left(\bm{\mu}^n,\bm{\Sigma}^{n,n}\right),
\]
to obtain
\begin{eqnarray}\label{eq:Advection_posterior}
&&u^{n+1}(x^{n+1}_*)\ | \left[\begin{array}{c}
u^{n+1}(1) - u^{n+1}(0) =0 \\ 
u^{n+\tau_2}(1) - u^{n+\tau_2}(0) =0 \\ 
u^{n+\tau_1}(1) - u^{n+\tau_1}(0) =0
\end{array} \right]
\\ 
&&\sim \mathcal{N}\left(\mu^{n+1}(x_*^{n+1}),\Sigma^{n+1,n+1}(x_*^{n+1},x_*^{n+1})\right),\nonumber
\end{eqnarray}
where
\[
\mu^{n+1}(x_*^{n+1}) = \bm{q}^T \bm{K}^{-1}\left[\begin{array}{c}
0 \\ 
0 \\ 
0 \\ 
\bm{\mu}^n \\
\bm{\mu}^n \\
\bm{\mu}^{n}
\end{array} \right],
\]
and
\begin{eqnarray*}
&&\Sigma^{n+1,n+1}(x_*^{n+1},x_*^{n+1}) = k^{n+1,n+1}_{u,u}(x^{n+1}_*,x^{n+1}_*) - \bm{q}^T\bm{K}^{-1}\bm{q} \\
&& + \bm{q}^T\bm{K}^{-1} \left[\begin{array}{cccccc}
 0 & 0 & 0 & 0 & 0 & 0\\ 
  & 0 & 0 & 0 & 0 & 0\\ 
  &  & 0 & 0 & 0 & 0\\ 
  &  &  & \bm{\Sigma}^{n,n} & \bm{\Sigma}^{n,n} & \bm{\Sigma}^{n,n}\\
  &  &  &  & \bm{\Sigma}^{n,n} & \bm{\Sigma}^{n,n} \\
  &  &  &  &  & \bm{\Sigma}^{n,n}
\end{array} \right]\bm{K}^{-1}\bm{q}.
\end{eqnarray*}
Now, we can use the resulting posterior distribution (\ref{eq:Advection_posterior}) to obtain the artificially generated data $\{\bm{x}^{n+1},\bm{u}^{n+1}\}$ with
\begin{equation}\label{eq:Advection_artificial_data}
\bm{u}^{n+1} \sim \mathcal{N}\left(\bm{\mu}^{n+1},\bm{\Sigma}^{n+1,n+1}\right).
\end{equation}

\subsection{Heat equation}\label{Appendix:Heat}
\subsubsection{Prior}
The covariance functions for the Heat equation are given by
\begin{eqnarray}
k^{n+1,n+1}_{u,v} &=& \frac{d}{d x_2'} k^{n+1,n+1}_{u,u},\\ 
k^{n+1,n}_{u,3} &=& k^{n+1,n+1}_{u,u} - \frac{1}{2}\Delta t \frac{d^2}{d x_1'^2}k^{n+1,n+1}_{u,u} - \frac{1}{2}\Delta t \frac{d^2}{d x_2'^2}k^{n+1,n+1}_{u,u},\nonumber\\
k^{n+1,n+1}_{v,v} &=& \frac{d}{d x_2}\frac{d}{d x_2'} k^{n+1,n+1}_{u,u},\nonumber\\
k^{n+1,n}_{v,3} &=& \frac{d}{d x_2}k^{n+1,n+1}_{u,u} - \frac{1}{2}\Delta t \frac{d}{d x_2}\frac{d^2}{d x_1'^2}k^{n+1,n+1}_{u,u} - \frac{1}{2}\Delta t \frac{d}{d x_2}\frac{d^2}{d x_2'^2}k^{n+1,n+1}_{u,u},\nonumber\\
k^{n,n}_{u,v} &=& \frac{d}{d x_2'} k^{n,n}_{u,u},\nonumber\\
k^{n,n}_{u,3} &=& - \frac{1}{2}\Delta t \frac{d^2}{d x_1'^2}k^{n,n}_{u,u} - \frac{1}{2}\Delta t \frac{d^2}{d x_2'^2}k^{n,n}_{u,u},\nonumber\\
k^{n,n}_{u,1} &=& k^{n,n}_{u,u}\nonumber\\
k^{n,n}_{v,v} &=& \frac{d}{d x_2}\frac{d}{d x_2'} k^{n,n}_{u,u},\nonumber\\
k^{n,n}_{v,3} &=& - \frac{1}{2}\Delta t \frac{d}{d x_2}\frac{d^2}{d x_1'^2}k^{n,n}_{u,u} - \frac{1}{2}\Delta t \frac{d}{d x_2}\frac{d^2}{d x_2'^2}k^{n,n}_{u,u},\nonumber\\
k^{n,n}_{v,1} &=& \frac{d}{d x_2} k^{n,n}_{u,u},\nonumber
\end{eqnarray}
and
\begin{eqnarray}
k^{n,n}_{3,3} &=& k^{n+1,n+1}_{u,u} - \frac{1}{2}\Delta t \frac{d^2}{d x_1'^2}k^{n+1,n+1}_{u,u} - \frac{1}{2}\Delta t \frac{d^2}{d x_2'^2}k^{n+1,n+1}_{u,u}\\
&+&\frac{1}{4} \Delta t^2 \frac{d^2}{d x_1^2}\frac{d^2}{d x_1'^2}k^{n,n}_{u,u} +\frac{1}{4} \Delta t^2 \frac{d^2}{d x_1^2}\frac{d^2}{d x_2'^2}k^{n,n}_{u,u}\nonumber\\ 
&-& \frac12 \Delta t \frac{d^2}{d x_1^2} k^{n+1,n+1}_{u,u} + \frac14 \Delta t^2 \frac{d^2}{d x_1^2} \frac{d^2}{d x_1'^2} k^{n+1,n+1}_{u,u} + \frac14 \Delta t^2 \frac{d^2}{d x_1^2} \frac{d^2}{d x_2'^2} k^{n+1,n+1}_{u,u}\nonumber\\
&+&\frac{1}{4} \Delta t^2 \frac{d^2}{d x_2^2}\frac{d^2}{d x_1'^2}k^{n,n}_{u,u} +\frac{1}{4} \Delta t^2 \frac{d^2}{d x_2^2}\frac{d^2}{d x_2'^2}k^{n,n}_{u,u}\nonumber\\
&-& \frac12 \Delta t \frac{d^2}{d x_2^2} k^{n+1,n+1}_{u,u} + \frac14 \Delta t^2 \frac{d^2}{d x_2^2} \frac{d^2}{d x_1'^2} k^{n+1,n+1}_{u,u} + \frac14 \Delta t^2 \frac{d^2}{d x_2^2} \frac{d^2}{d x_2'^2} k^{n+1,n+1}_{u,u},\nonumber
\end{eqnarray}
\begin{eqnarray*}
k^{n,n}_{3,1} &=& - \frac{1}{2}\Delta t \frac{d^2}{d x_1^2}k^{n,n}_{u,u} - \frac{1}{2}\Delta t \frac{d^2}{d x_2^2}k^{n,n}_{u,u},\nonumber\\
k^{n,n}_{1,1} &=& k^{n,n}_{u,u}.\nonumber
\end{eqnarray*}

\subsubsection{Training}
The matrix $\bm{K}$ used in the distribution (\ref{eq:Heat_NLML}) is given by
\begin{eqnarray*}
\bm{K} = \left[\begin{array}{cccccc}
\bm{K}^{n+1,n+1}_{D,D} & \bm{K}^{n+1,n+1}_{D,N} & 0 & 0 & \bm{K}^{n+1,n}_{D,3} & 0\\
 & \bm{K}^{n+1,n+1}_{N,N} & 0 & 0 & \bm{K}^{n+1,n}_{N,3} & 0\\
 & & \bm{K}^{n,n}_{D,D} & \bm{K}^{n,n}_{D,N} & \bm{K}^{n,n}_{D,3} & \bm{K}^{n,n}_{D,1}\\
 & & & \bm{K}^{n,n}_{N,N} & \bm{K}^{n,n}_{N,3} & \bm{K}^{n,n}_{N,1}\\ 
 & & & & \bm{K}^{n,n}_{3,3} & \bm{K}^{n,n}_{3,1}\\ 
 & & & & & \bm{K}^{n,n}_{1,1}\\ 
\end{array}\right].
\end{eqnarray*}
Here,
\begin{eqnarray}
\bm{K}^{n+1,n+1}_{D,D} &=& k^{n+1,n+1}_{u,u}\left((\bm{x}_{1,D}^{n+1},\bm{x}_{2,D}^{n+1}),(\bm{x}_{1,D}^{n+1},\bm{x}_{2,D}^{n+1})\right) + \sigma^2_{D,n+1} I,\\
\bm{K}^{n+1,n+1}_{D,N} &=& k^{n+1,n+1}_{u,v}\left((\bm{x}_{1,D}^{n+1},\bm{x}_{2,D}^{n+1}),(\bm{x}_{1,N}^{n+1},\bm{x}_{2,N}^{n+1})\right),\nonumber\\
\bm{K}^{n+1,n}_{D,3} &=& k^{n+1,n}_{u,3}\left((\bm{x}_{1,D}^{n+1},\bm{x}_{2,D}^{n+1}),(\bm{x}_{1}^n,\bm{x}_{2}^n)\right),\nonumber\\
\bm{K}^{n+1,n+1}_{N,N} &=& k^{n+1,n+1}_{v,v}\left((\bm{x}_{1,N}^{n+1},\bm{x}_{2,N}^{n+1}),(\bm{x}_{1,N}^{n+1},\bm{x}_{2,N}^{n+1})\right) + \sigma^2_{N,n+1} I,\nonumber\\
\bm{K}^{n+1,n}_{N,3} &=& k^{n+1,n}_{v,3}\left((\bm{x}_{1,N}^{n+1},\bm{x}_{2,N}^{n+1}),(\bm{x}_1^n,\bm{x}_2^n)\right),\nonumber
\end{eqnarray}
\begin{eqnarray}
\bm{K}^{n,n}_{D,D} &=& k^{n,n}_{u,u}\left((\bm{x}_{1,D}^{n},\bm{x}_{2,D}^{n}),(\bm{x}_{1,D}^{n},\bm{x}_{2,D}^{n})\right) + \sigma^2_{D,n} I,\nonumber\\
\bm{K}^{n,n}_{D,N} &=& k^{n,n}_{u,v}\left((\bm{x}_{1,D}^{n},\bm{x}_{2,D}^{n}),(\bm{x}_{1,N}^{n},\bm{x}_{2,N}^{n})\right),\nonumber\\
\bm{K}^{n,n}_{D,3} &=& k^{n,n}_{u,3}\left((\bm{x}_{1,D}^{n},\bm{x}_{2,D}^{n}),(\bm{x}_{1}^n,\bm{x}_{2}^n)\right),\nonumber\\
\bm{K}^{n,n}_{D,1} &=& k^{n,n}_{u,1}\left((\bm{x}_{1,D}^{n},\bm{x}_{2,D}^{n}),(\bm{x}_{1}^n,\bm{x}_{2}^n)\right),\nonumber\\
\bm{K}^{n,n}_{N,N} &=& k^{n,n}_{v,v}\left((\bm{x}_{1,N}^{n},\bm{x}_{2,N}^{n}),(\bm{x}_{1,N}^{n},\bm{x}_{2,N}^{n})\right) + \sigma^2_{N,n} I,\nonumber\\
\bm{K}^{n,n}_{N,3} &=& k^{n,n}_{v,3}\left((\bm{x}_{1,N}^{n},\bm{x}_{2,N}^{n}),(\bm{x}_1^n,\bm{x}_2^n)\right),\nonumber\\
\bm{K}^{n,n}_{N,1} &=& k^{n,n}_{v,1}\left((\bm{x}_{1,N}^{n},\bm{x}_{2,N}^{n}),(\bm{x}_1^n,\bm{x}_2^n)\right),\nonumber
\end{eqnarray}
\begin{eqnarray}
\bm{K}^{n,n}_{3,3} &=& k^{n,n}_{3,3}\left((\bm{x}_{1}^{n},\bm{x}_{2}^{n}),(\bm{x}_1^n,\bm{x}_2^n)\right) + \sigma^2_{n} I,\nonumber\\
\bm{K}^{n,n}_{3,1} &=& k^{n,n}_{3,1}\left((\bm{x}_{1}^{n},\bm{x}_{2}^{n}),(\bm{x}_1^n,\bm{x}_2^n)\right),\nonumber\\
\bm{K}^{n,n}_{1,1} &=& k^{n,n}_{1,1}\left((\bm{x}_{1}^{n},\bm{x}_{2}^{n}),(\bm{x}_1^n,\bm{x}_2^n)\right) + \sigma^2_{n} I.\nonumber
\end{eqnarray}

\subsubsection{Posterior}
In order to predict $u^{n+1}(x^{n+1}_{1,*},x^{n+1}_{2,*})$ at a new test point $(x^{n+1}_{1,*},x^{n+1}_{2,*})$, we use
\begin{eqnarray*}
&&u^{n+1}(x^{n+1}_{1,*},x^{n+1}_{2,*})\ | \left[\begin{array}{c}
\bm{u}^{n+1}_{D}\\
\bm{v}^{n+1}_{N}\\
\bm{u}^{n}_{D}\\
\bm{v}^{n}_{N}\\
\bm{u}^n\\
\bm{u}^n
\end{array} \right] \sim \\
&&\mathcal{N}\left(
\bm{q}^T \bm{K}^{-1}\left[\begin{array}{c}
\bm{u}^{n+1}_{D}\\
\bm{v}^{n+1}_{N}\\
\bm{u}^{n}_{D}\\
\bm{v}^{n}_{N}\\
\bm{u}^n\\
\bm{u}^n
\end{array} \right],k^{n+1,n+1}_{u,u}\left((x^{n+1}_{1,*},x^{n+1}_{2,*}),(x^{n+1}_{1,*},x^{n+1}_{2,*})\right) - \bm{q}^T\bm{K}^{-1}\bm{q}\right),
\end{eqnarray*}
where
\[
\bm{q} := \left[\begin{array}{c}
k^{n+1,n+1}_{u,u}\left((\bm{x}^{n+1}_{1,D},\bm{x}^{n+1}_{2,D}),(x^{n+1}_{1,*},x^{n+1}_{2,*})\right)\\
k^{n+1,n+1}_{v,u}\left((\bm{x}^{n+1}_{1,N},\bm{x}^{n+1}_{2,N}),(x^{n+1}_{1,*},x^{n+1}_{2,*})\right)\\
0\\
0\\
k^{n,n+1}_{3,u}\left((\bm{x}^n_1,\bm{x}^n_2),(x^{n+1}_{1,*},x^{n+1}_{2,*})\right)\\ 
0
\end{array} \right].
\]

\subsubsection{Propagating Uncertainty}
To propagate the uncertainty associate with the noisy initial data through time we have to marginalize out the artificially generated data $\{(\bm{x}^n_1,\bm{x}^n_2), \bm{u}^n\}$ by employing
\[
\bm{u}^n \sim \mathcal{N}\left(\bm{\mu}^n,\bm{\Sigma}^{n,n}\right),
\]
to obtain
\begin{eqnarray}\label{eq:Heat_posterior}
&&u^{n+1}(x^{n+1}_{1,*},x^{n+1}_{2,*})\ | \left[\begin{array}{c}
\bm{u}^{n+1}_{D}\\
\bm{v}^{n+1}_{N}\\
\bm{u}^{n}_{D}\\
\bm{v}^{n}_{N}\\
\end{array} \right]
\\ 
&&\sim \mathcal{N}\left(\mu^{n+1}(x^{n+1}_{1,*},x^{n+1}_{2,*}),\Sigma^{n+1,n+1}((x^{n+1}_{1,*},x^{n+1}_{2,*}),(x^{n+1}_{1,*},x^{n+1}_{2,*}))\right),\nonumber
\end{eqnarray}
where
\[
\mu^{n+1}(x^{n+1}_{1,*},x^{n+1}_{2,*}) = \bm{q}^T \bm{K}^{-1}\left[\begin{array}{c}
\bm{u}^{n+1}_{D}\\
\bm{v}^{n+1}_{N}\\
\bm{u}^{n}_{D}\\
\bm{v}^{n}_{N}\\
\bm{\mu}^{n}\\
\bm{\mu}^{n}
\end{array} \right],
\]
and
\begin{eqnarray*}
&&\Sigma^{n+1,n+1}(x_*^{n+1},x_*^{n+1}) = k^{n+1,n+1}_{u,u}(x^{n+1}_*,x^{n+1}_*) - \bm{q}^T\bm{K}^{-1}\bm{q} \\
&& + \bm{q}^T\bm{K}^{-1} \left[\begin{array}{cccccc}
 0 & 0 & 0\\
  & \bm{\Sigma}^{n,n} &\bm{\Sigma}^{n,n}\\
  & &\bm{\Sigma}^{n,n}  
\end{array} \right]\bm{K}^{-1}\bm{q}.
\end{eqnarray*}
Now, we can use the resulting posterior distribution (\ref{eq:Heat_posterior}) to obtain the artificially generated data $\{(\bm{x}^{n+1}_1,\bm{x}^{n+1}_2),\bm{u}^{n+1}\}$ with
\begin{equation}\label{eq:Heat_artificial_data}
\bm{u}^{n+1} \sim \mathcal{N}\left(\bm{\mu}^{n+1},\bm{\Sigma}^{n+1,n+1}\right).
\end{equation}

\end{document}